\documentclass[10pt,journal,compsoc]{IEEEtran}
\ifCLASSINFOpdf
\else
\fi
\usepackage{amsmath}
\usepackage{amssymb}
\usepackage{mathrsfs}
\usepackage{graphicx}
\usepackage{bm}
\usepackage{wrapfig}
\usepackage{amsmath}
\usepackage{amssymb}
\usepackage{algorithmic}
\usepackage{algorithm}
\usepackage{subcaption}
\usepackage[switch]{lineno}
\usepackage{booktabs} 
\usepackage{color}
\usepackage{colortbl}
\usepackage{url}
\usepackage{xcolor}
\usepackage{multirow}
\usepackage{epstopdf}
\newcommand{\EE}{\mathbb{E}}

\newcommand{\CC}{\mathcal{C}}

\newcommand{\Tc}{\mathcal{T}}
\newcommand{\RR}{\mathbb{R}}
\newcommand{\SI}{\textrm{Sign}}

\newtheorem{lemma}{Lemma}

\newtheorem{theorem}{Theorem}

\newtheorem{assumption}{Assumption}
\newtheorem{proposition}{Proposition}
\begin{document}

 \title{Rethinking SIGN Training: Provable Nonconvex Acceleration without   First- and Second-Order   Gradient Lipschitz}

\author{Tao Sun, Congliang Chen, Peng Qiao, Li Shen, Xinwang Liu,
        Dongsheng Li
\thanks{

Tao Sun, Peng Qiao, Xinwang Liu, and Dongsheng Li are with the
College of Computer, National University of Defense Technology,
Changsha, 410073, Hunan,  China (e-mails: \texttt{suntao.saltfish@outlook.com,pengqiao@nudt.edu.cn}, \texttt{xinwangliu@nudt.edu.cn,dsli@nudt.edu.cn}).

Congliang Chen is with The Chinese University of Hong Kong, Shenzhen, China (e-mail: \texttt{congliangchen@link.cuhk.edu.cn}).

Li Shen is with  JD Explore Academy, Beijing, China
  (e-mail: \texttt{mathshenli@gmail.com}).

}
}

\markboth{Journal of \LaTeX\ Class Files,~Vol.~14, No.~8, August~2015}%
{Shell \MakeLowercase{\textit{et al.}}: Bare Demo of IEEEtran.cls for IEEE Journals}

\maketitle

\begin{abstract}
 Sign-based stochastic methods have gained attention due to their ability to achieve robust performance despite using only the sign information for parameter updates. However, the current convergence analysis of sign-based methods relies on the strong assumptions of first-order gradient Lipschitz and second-order gradient Lipschitz, which may not hold in practical tasks like deep neural network training that involve high non-smoothness.
In this paper, we revisit sign-based methods and analyze their convergence under more realistic assumptions of first- and second-order smoothness. We first establish the convergence of the sign-based method under weak first-order Lipschitz. Motivated by the weak first-order Lipschitz, we propose a relaxed second-order condition that still allows for nonconvex acceleration in sign-based methods.
Based on our theoretical results, we gain insights into the computational advantages of the recently developed LION algorithm. In distributed settings, we prove that this nonconvex acceleration persists with linear speedup in the number of nodes, when utilizing fast communication compression gossip protocols.
The novelty of our theoretical results lies in that they are derived under much weaker assumptions, thereby expanding the provable applicability of sign-based algorithms to a wider range of problems.

\end{abstract}


\section{Introduction}

The widely used Stochastic Gradient Descent (SGD) \cite{robbins1951stochastic} is the main optimization algorithm for solving the fundamental optimization problem arising in machine learning and statistics
\begin{align}\label{model}
\min_{\bm{w}\in\RR^d} f(\bm{w}):=\EE_{\xi\sim\mathcal{D}} f(\bm{w};\xi),
\end{align}
Here, $\mathcal{D}$ denotes the probability distribution over the statistical sample space $\Xi$. However, vanilla SGD may not be very efficient in certain cases, such as communication-efficient training. In such scenarios, methods with compressed stochastic gradients are often employed. One elegant method is to employ the sign of each coordinate of the stochastic gradient, known as SignSGD. In each iteration of SignSGD \cite{pmlr-v80-bernstein18a}, a data point $\xi^t$ is sampled independently from the distribution $\mathcal{D}$, and the parameter update is performed as follows
\begin{align}\label{alg}
{\bm w}^{t+1}={\bm w}^{t}-\gamma \SI[\nabla f(\bm{w}^t;\xi^t)],
\end{align}
where $\gamma > 0$ denotes the step size.
However, \eqref{alg} is not convergent when used with a small batch size \cite{pmlr-v80-bernstein18a,bernstein2018signsgd}. To address this issue, several techniques such as error feedback and momentum have been developed to improve the convergence of SignSGD \cite{pmlr-v97-karimireddy19a,momentum2023}.
Despite these improvements, SignSGD still discards a significant amount of information in each iteration, resulting in worse performance compared to SGD from the intuitive perspective. However, surprising numerical results have shown that SignSGD can be comparable to and even faster than SGD in some cases \cite{pmlr-v80-bernstein18a,bernstein2018signsgd,pmlr-v97-karimireddy19a,momentum2023}.
Indeed, in the paper by \cite{chen2023symbolic}, the authors propose a sign-based method called LION that demonstrates adorable
 numerical performance for training and effectively alleviates overfitting.
Although a provably accelerated algorithm based on sign is proposed in \cite{momentum2023}, the proofs rely on a strong smoothness assumption: the gradient needs to follow first-order and second-order Lipschitz conditions, which are both unrealistic for training deep neural networks. As a result, the theory presented in \cite{momentum2023} cannot explain the advantage of sign-based methods.
Indeed, the current convergence results of SignSGD are based on the assumption of first-order gradient Lipschitzness \cite{pmlr-v80-bernstein18a,bernstein2018signsgd,pmlr-v97-karimireddy19a,momentum2023}. However, this assumption cannot adequately explain why SignSGD exhibits convergence in deep neural network training, let alone the nonconvex acceleration.
Therefore, the main problem we aim to address in this paper is
\begin{center}
\textit{Is the convergence of sign-based methods possible without relying on the assumption of first-order gradient Lipschitzness? Can we establish the theoretical advantages of sign-based methods without assuming the first- and second-order gradient Lipschitz conditions?}
\end{center}
In this paper, we provide an affirmative answer to this question.

\subsection{First- and Second-Order Gradient Lipschitz}
The gradient first-order Lipschitz smoothness is a crucial assumption for the convergence of nonconvex SGD.
Mathematically,   it usually assumes that   the objective function $f$ satisfies
\begin{equation}\label{Lips}
\|\nabla f(\bm{x})-\nabla f(\bm{y})\|\leq L\|\bm{x}-\bm{y}\|\tag{first-order Lipschitz}
\end{equation}
for constant $L>0$ and any $\bm{x},\bm{y}\in\RR^d$. Although \ref{Lips}  is commonly used in the analysis of stochastic optimization \cite{bottou2018optimization}, it cannot hold in numerous machine learning tasks, especially in the neural network training \cite{zhang2019gradient}. The \ref{Lips} does not even hold for the following simple one-rank matrix approximation \cite{markovsky2012low}
\begin{equation}\label{se}
 \min_{\bm{x}\in\RR^{d}} \{D(\bm{x}):=\frac{1}{2}\|\bm{x}\bm{x}^{\top}-\bm{Y}\|_{F}^{2}\},
\end{equation}
where $\bm{Y}\in\RR^{d\times d}$.
If function $D$ satisfies the \ref{Lips}, it must follow
$\|\nabla^2 D(\bm{x})\|_{\textrm{op}}<+\infty$ for any $\bm{x}\in\RR^{d}$.
However, letting $\bm{x}=t\bm{e}_{1}$ with $\bm{e}_{1}\in\RR^{m\times d}$ being almost zero vector whose only the  first element is 1, we have
$\|\nabla^2 D(\bm{x})\|_{\textrm{op}}=\Big\|\|\bm{x}\|^2\cdot \mathbb{I}+2\bm{x}\bm{x}^{\top}-\bm{Y}\Big\|_{\textrm{op}}\geq \Big|\|t^2\cdot \mathbb{I}+2t^2\bm{e}_1\bm{e}_1^{\top}\|_{\textrm{op}}-\|\bm{Y}\|_{\textrm{op}}\Big|\geq 3t^2-\|\bm{Y}\|_{\textrm{op}}\rightarrow+\infty$ as $t\rightarrow+\infty$, indicating that $D$ does not obey the \ref{Lips}.
Hence, the theory built on   \ref{Lips} is indeed hard to explain the convergence phenomenon of various optimization problems that fail to obey the \ref{Lips}.

To bridge the gap between the convergence and the untenable \ref{Lips} in neural networks training, \cite{zhang2019gradient,zhang2020improved} propose a more realistic first-order assumption for the gradient.
\begin{assumption}\label{ass1}
There exist   constants $L_1,r>0$ and $L_2\geq0$, for any  $\bm{x},\bm{y}\in\RR^d$ such that $\|{\bm y}-{\bm x}\|\leq r$,
function $f$ obeys
\begin{equation}\label{newsmooth}
\|\nabla f({\bm y})-\nabla  f({\bm x})\|\leq (L_1+L_2\|\nabla  f({\bm x})\|)\|{\bm y}-{\bm x}\|.
\end{equation}
\end{assumption}
It is clear that  \eqref{newsmooth}  reduces to \ref{Lips} as $L_2=0$.
In paper \cite{zhang2019gradient}, the authors   numerically verified that
Assumption \ref{ass1} can hold in plenty of tasks. The function in \eqref{se} also follows Assumption \ref{ass1} and details could be found in the supplementary materials.
It has been proved that the SGD may not be convergent under Assumption \ref{ass1} because the difference between two gradients may be unbounded
even if two points are very close. To guarantee the convergence of the SGD with the weak Lipschitz property, extra operations on the gradient are necessary, for example,
the gradient clipping \cite{pascanu2013difficulty,goodfellow2016deep,merity2018regularizing,gehring2017convolutional,menon2019can}. Although the introduction of gradient clipping promises the convergence under weak first-order smoothness, it requires a stronger assumption on the noise compared to SGD, specifically, the almost surely (a.s.) bounded noise \cite{zhang2019gradient}.

There has been a line of research on accelerating the nonconvex SGD.
The nonconvex accelerators require the objective functions to be extra second-order Lipschitz smoothness besides \ref{Lips} to get the theoretical acceleration because the convexity is unavailable \cite{carmon2017convex,carmon2018accelerated,carmon2020first,agarwal2016finding,jin2018accelerated,li2022restarted}, i.e., it is assumed
\begin{equation}\label{Lips2}
\|\nabla^2 f(\bm{x})-\nabla^2 f(\bm{y})\|_{\textrm{op}}\leq H\|\bm{x}-\bm{y}\| \tag{second-order Lipschitz}
\end{equation}
for some constant $H>0$.
Nevertheless, the first-order Lipschitz is absent in many applications, let alone the second-order Lipschitz property.
For example, consider the simple function \eqref{se}, which cannot obey \ref{Lips2} either. Indeed, if \ref{Lips2} holds, the $\nabla^3 f(\bm{x})$ must be uniformly bounded. However, with direct computations,
$|[\nabla^3 f(\bm{x})]_{1,1,1}|=4|\bm{x}_1|\rightarrow+\infty$ as $|\bm{x}_1|\rightarrow+\infty$.

\subsection{More Related Works}
We briefly review three kinds of related works:  weak smoothness,  accelerated nonconvex algorithms, and sign-based stochastic methods.

\textbf{Weak  smoothness.}
Investigating the stochastic optimization algorithms under weak smoothness is necessary and meaningful because \ref{Lips}   always fails to hold in practice.
Zhang et al.  first consider the  smoothness
 \begin{equation}\label{tempsmooth}
   \|\nabla^2 f(\bm{x})\|_{\textrm{op}}\leq L_1+L_2\|\nabla  f({\bm x})\|
 \end{equation}
 for the clipping SGD, and prove that the clipping SGD exhibits superior performance when the norms of the gradients are increasing rapidly \cite{zhanggradient}.
However, it is important to note that condition \eqref{tempsmooth} is not a direct relaxation of the Lipschitz condition \ref{Lips} due to its requirement of twice differentiability. In the subsequent work \cite{zhang2020improved}, the authors established that under the assumption of twice differentiability, condition \eqref{tempsmooth} represents a weaker form of smoothness, denoted as \eqref{newsmooth}, which does not necessitate twice differentiability and is thus considerably more flexible than the \ref{Lips}.
Furthermore, the authors in \cite{zhang2020improved} derived the convergence rate of a general clipping SGD algorithm under the weak smoothness condition \eqref{newsmooth}. This analysis contributes to our understanding of the behavior and effectiveness of stochastic optimization algorithms in scenarios where the Lipschitz condition \ref{Lips} cannot be satisfied.
It is noteworthy that the analyses presented in the aforementioned papers incorporate the additional assumption of almost-sure boundedness on the noise, which is stronger in comparison to the commonly used assumption of bounded variance \cite{bottou2018optimization}. This stronger assumption enables a more rigorous analysis and comprehension of stochastic optimization algorithms under weak smoothness.
In \cite{jin2021non}, the authors prove the convergence of normalized SGD under weak first-order Lipschitz smoothness, which further relaxes the assumption of almost-sure bounded noise used in the clipping method.
Besides convergence analysis, the generalization property under weak smoothness has also received considerable attention \cite{mai2021stability} because  evaluating the testing accuracy of a machine learning model holds significant importance for both researchers and practitioners.
Furthermore, in the study conducted in \cite{sun2022note}, the authors investigate a probability distribution sampling algorithm that samples from a given distribution under the weak smoothness assumption. The analysis of this algorithm sheds light on the behavior and performance of sampling-based optimization algorithms in scenarios where weak smoothness is present.

\textbf{Accelerated nonconvex optimization.} 
Extensive research has been conducted on accelerating first-order optimization algorithms for convex deterministic cases, such as Nesterov's method \cite{nesterov1983method}. However, it has been found in \cite{assran2020convergence} that Nesterov's method may diverge even for strongly convex and twice continuously differentiable functions in stochastic optimization, when using the usual choice of step size and momentum. This poses a challenge in extending Nesterov's method to the general nonconvex stochastic case.
Interestingly, there exist provable deterministic nonconvex accelerated optimization algorithms, but they require the function to exhibit second-order Lipschitz smoothness due to the loss of convexity. Initially, deterministic nonconvex acceleration approaches utilized nested-loop schemes, where each iteration involved solving a sub-optimization problem \cite{carmon2017convex,carmon2018accelerated,carmon2020first,agarwal2016finding}. However, these nested algorithms are complex and involve multiple hyperparameters, making them impractical to implement.
To address this, several single-loop nonconvex accelerated schemes have been developed \cite{jin2018accelerated,li2022restarted,cutkosky2020momentum}. These methods have a similar complexity of $\mathcal{O}(T^{-\frac{4}{7}})$ for finding the first-order stationary point.


In the stochastic setting, the convergence of nonconvex accelerated algorithms is generally degraded to $\mathcal{O}(T^{-\frac{2}{7}})$ due to the presence of noise \cite{tripuraneni2018stochastic,allen2018natasha,fang2019sharp,zhou2018finding,cutkosky2020momentum}. These algorithms still require the first- and second-order Lipschitz smoothness properties for convergence guarantees.
In \cite{tripuraneni2018stochastic}, a nonconvex accelerated stochastic algorithm based on the cubic-regularized Newton method is proposed. However, this method requires solving a quadratic problem in each iteration, which can be computationally expensive.
In papers \cite{allen2018natasha,zhou2018finding}, the authors bypass the need for iterative sub-optimization by incorporating variance reduction techniques while maintaining the same convergence rate. This saves computational resources and makes the algorithms more practical.
In \cite{fang2019sharp}, the effectiveness of a ball-controlled mechanism as a restart criterion for achieving acceleration is proven. This criterion allows the algorithm to effectively escape poor local optima.
Alternatively, in \cite{cutkosky2020momentum}, the authors present a simple nonconvex accelerated method that combines gradient normalization and momentum techniques. This method does not require sub-minimization, variance reduction, or restart, making it easier to implement.
Overall, these nonconvex accelerated algorithms provide promising approaches for improving the convergence rates of nonconvex stochastic optimization problems under very  smooth assumptions.


\textbf{Sign-based stochastic methods.}

There has been extensive research conducted to understand and improve sign-based methods.
For instance, a theoretical analysis of the iteration error bound on SignSGD is provided in \cite{pmlr-v80-bernstein18a}. A variant of SignSGD called major vote is proposed in \cite{bernstein2018signsgd}, which is specifically designed for distributed settings and only requires transmitting the sign information.
In the context of robust deep learning and black-box adversarial attacks, \cite{liu2018signsgd} propose zeroth order SignSGD, which eliminates the need for employing the stochastic gradient directly.
Exploiting a sign-based gradient estimation approach, \cite{Al-Dujaili2020Sign} present a novel black-box adversarial attack algorithm.
In the paper \cite{NEURIPS2020_a7f0d2b9}, the authors leverage the SignSGD algorithm to develop a coding method that minimizes the communication load between workers and a master node while guaranteeing Byzantine-robustness for distributed learning.
Similarly, in \cite{2020Stochastic}, a sign-based method, specifically SignSGD, is applied to federated training tasks. The study proves convergence guarantees for this approach, demonstrating its effectiveness in federated learning scenarios.
However, it is worth noting that most of the aforementioned SignSGD papers require an increasing batch size to ensure their convergence. More recently, some researchers have proposed modifications to SignSGD that aim to reduce the sampling costs associated with it.
For instance, in the paper \cite{pmlr-v97-karimireddy19a}, the authors introduce the error feedback technique to eliminate the need for the large sampling assumption in SignSGD.
In another study, \cite{pmlr-v139-safaryan21a}, the scheme of SignSGD is modified by comparing global objective function values in each iteration, resulting in a different variant of SignSGD.
By considering a coordinate Lipschitz-like property, a robust general SignSGD algorithm with stepsize adjusted by historical gradients is proposed in \cite{crawshaw2022robustness}.
In a more recent paper \cite{momentum2023}, the authors prove that simple momentum alone is sufficient to guarantee the convergence of SignSGD, even under weaker assumptions.
Moreover, the authors of \cite{momentum2023} propose an accelerated version of SignSGD with theoretical guarantees under first- and second-order smoothness.
In the paper \cite{chen2023symbolic}, the authors propose a sign-based method called LION, which demonstrates substantially faster performance compared to the popular Adam optimizer \cite{DBLP:journals/corr/KingmaB14} on vision Transformers and diffusion models.
Numerics has shown the efficiency of the LION but with few theoretical explanations.
\subsection{Contributions}
This paper aims to demonstrate the acceleration of accelerated Sign methods under weak smoothness assumptions, providing a theoretical understanding of their advantages in practical training. The main contributions of this work can be summarized as follows:

\begin{itemize}
\item We present a proof showcasing the convergence of the sign-based method, assuming weak first-order gradient smoothness. Our findings indicate that by utilizing simple momentum, the sign-based method can achieve convergence rates equivalent to those of SGD.

   \item We propose a more realistic assumption to characterize the Hessian property, which is significantly weaker than the second-order Lipschitz smoothness. Surprisingly, we are still able to demonstrate nonconvex acceleration  of the accelerated SignSGD with this weak first- and second-order Lipschitz smoothness, which greatly expands the potential applications. Furthermore, we provide an explanation as to why accelerated SignSGD accelerations can outperform other nonconvex accelerated methods when gradients are exploding.

  \item  We show that the accelerated SignSGD exhibits a close relationship with the recently popular LION algorithm \cite{chen2023symbolic}. Theoretical analysis demonstrates that nonconvex convergence and acceleration are still achievable under weaker assumptions, shedding light on some advantages of LION.

  \item We apply Sign methods to nonconvex communication-efficient distributed settings and develop a novel algorithm. This new distributed algorithm achieves convergence under the assumption of weak first-order smoothness and is further proven to be faster in terms of both iterations and communications under the assumption of weak second-order smoothness.

\end{itemize}

\noindent\textbf{Notation:} 
This paper uses  $\EE[\cdot]$ to denote the expectation with respect to the underlying probability space. We  use  $\|\cdot\|$ to denote the $L_2$-norm of a vector and  $\|\cdot\|_{\textrm{op}}$ to denote the spectral norm of a matrix. We use $\nabla^3 f$ denote the third-order gradient of function $f$, i.e.,
$\nabla^3 f(\bm{x})\in\RR^{d^3}$ and $[\nabla^3 f(\bm{x})]_{i,j,k}=\frac{\partial f(\bm{x})}{\partial\bm{x}_i\partial\bm{x}_j\partial\bm{x}_k}$. Given an integer $n$ , we denote the set $[n]:=\{1,2,\ldots,n\}$.
The minimum value of the function $f$  is denoted as $\min f$. Given two sequences $\{a_t\}$ and $\{b_t\}$, we write $a_t=\mathcal{O}(b_t)$ if there exists  a positive constant $0<C<+\infty$ such that
$a_t \leq C b_t$, and write $a_t=\Theta(b_t)$ if $a_t=\mathcal{O}(b_t)$ and $b_t=\mathcal{O}(a_t)$.
Given variables $(\xi^0,\xi^1,\ldots,\xi^t)$, we denote the $\sigma$-field as $\chi^t:=\sigma(\xi^0,\xi^1,\ldots,\xi^t)$.

\section{Setup}
\subsection{More Realistic Second-Order Gradient Smoothness}
The \ref{Lips} assumption is quite strong for stochastic optimization, especially when it comes to the \ref{Lips2} assumption. Consequently, there have been numerous theoretical studies on accelerated (stochastic) nonconvex algorithms, but only a few of them have taken into account the practical aspects of training \cite{carmon2017convex,carmon2018accelerated,carmon2020first,agarwal2016finding,jin2018accelerated,li2022restarted,cutkosky2020momentum,tripuraneni2018stochastic,allen2018natasha,fang2019sharp,zhou2018finding,cutkosky2020momentum}.
To this end, this paper considers how to relax \ref{Lips2}.
Like the weak first-order gradient smoothness assumption \eqref{newsmooth},
we want to relax the \ref{Lips2}.
For simplicity, assume $f$ is a one-dimensional function. In this case, the \ref{Lips2} indeed indicates $g:=f'$ following the \ref{Lips}, i.e.,
$|g'(s)-g'(t)|\leq L|s-t|$ as $s,t\in\RR$. If $g$ follows the weak Lipschitz smoothness (Assumption \ref{ass1}), that is,
$|g'(s)-g'(t)|=\mathcal{O}[(1+|g'(t)|)|s-t|]=\mathcal{O}[(1+|f''(t)|)|s-t|]$. Furthermore, using \eqref{tempsmooth}, it follows
\begin{align*}
|f''(s)-f''(t)|&=|g'(s)-g'(t)|=\mathcal{O}[(1+|f'(t)|)|s-t|].
\end{align*}
Thus, we consider the \ref{Lips2} by replacing the fixed constant $H$ in \eqref{Lips2} with $H_1+H_2\|\nabla f({\bm x})\|$, and the new weak second-order gradient smoothness is presented as follows.
\begin{assumption}\label{ass2}
There exist   constants $H_1,R>0$ and $H_2\geq 0$, for  any  $\bm{x},\bm{y}\in\RR^d$ such that $\|{\bm y}-{\bm x}\|\leq R$
function $f$ obeys
\begin{equation}\label{newsmooth2}
\|\nabla^2 f({\bm y})-\nabla^2  f({\bm x})\|_{\textrm{op}}\leq (H_1+H_2\|\nabla  f({\bm x})\|)\|{\bm y}-{\bm x}\|.
\end{equation}
\end{assumption}
Assumption \ref{ass2} is the direct extension of the \ref{Lips2}, which is recovered  when $H_2=0,R=+\infty$ or
 the gradient $\|\nabla  f({\bm x})\|$ is uniformly bounded and $R=+\infty$.
 Assumption \ref{ass2} applies to the training problems that the \ref{Lips2} cannot hold, including the example \eqref{se}.
\begin{proposition}\label{pro1}
The objective function in problem \eqref{se} does not satisfy the \ref{Lips} and \ref{Lips2}, but obeys  Assumptions \ref{ass1} and \ref{ass2}.
\end{proposition}
In [Corollary A.4, \cite{zhang2020improved}], the authors proved that  Assumption \ref{ass1}   equals to
\eqref{tempsmooth} but with different constants if $f$ is twice differentiable.
Indeed for  Assumption \ref{ass2}, a similar equivalence hold when $\nabla^3 f$ exists.
\begin{proposition}\label{pro2}
If function $f$ has third-order gradient and Assumption \ref{ass1} holds, Assumption \ref{ass2} is equivalent to
$$\|\nabla^3 f(\bm{x})\|_{F}\leq \hat{H}_1\|\nabla  f({\bm x})\|+\hat{H}_2$$
for constants $\hat{H}_1,\hat{H}_2>0$.
\end{proposition}
With Proposition \ref{pro2}, we can see that besides example \eqref{se}, Assumption \ref{ass2} also holds for function $f(s)=s^n$ with $n>3$  and
$f(s)=\exp(s)$, which both fail to satisfy \ref{Lips2}.

\begin{figure}[H]
      \centering
\includegraphics[scale=0.6]{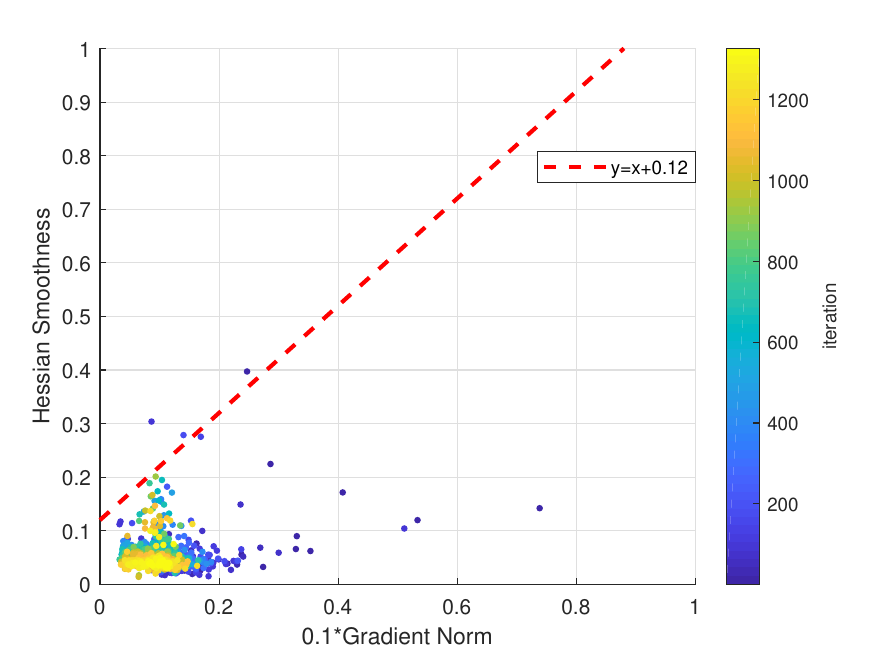}
        \caption{The figure above illustrates the relationship between the gradient norm and the local Hessian gradient Lipschitz constant during the training trajectory of AWD-LSTM on the PTB dataset. The colorbar represents the number of iterations completed during the training process.}\label{fig:third}
\end{figure}

\medskip
\textbf{Numerical Verification.}
Following the experiment reported in \cite{zhang2019gradient}, we conduct a similar experiment to investigate the gradient norm and local Hessian gradient Lipschitz constant on the training trajectory of AWD-LSTM \cite{merity2018regularizing} on PTB dataset \cite{mikolov2010recurrent} (see details in Appendix   \ref{sec2.1-numer}).
By analyzing Figure 3, it is evident that, for the majority of iterations (except for three iterations in the initial training phase), the Hessian smoothness is bounded by 0.1 times the gradient norm plus an offset of 0.12, as stated in Proposition 1. This observation supports the findings in Proposition 1 and further highlights the relationship between Hessian smoothness and the gradient norm. (Note: We use "0.1*Gradient Norm" as x-axis in Fig. \ref{fig:third}.)
\medskip

\textbf{Intuition for the  Phenomenon in  neural networks.} 
Consider the two-layer neural networks training with loss
$\frac{1}{2}\|{\bm W}_2\sigma({\bm W}_1{\bm a}_0)-\bm{y}\|$, where  ${\bm a}_0$ denotes the training samples, and ${\bm y}$ is the label, and $\sigma$ is an activation function with $|\sigma'(t)|<+\infty$ for any $t\in\RR$.
Given a very large penalty parameter $\bar{\rho}>0$,  an approximate problem is
\begin{equation}\label{penalty}
\begin{aligned}
&\min_{{\bm z},{\bm W}_1,{\bm W}_2} f({\bm z},{\bm W}_1,{\bm W}_2) \\
&\qquad\qquad:=\frac{1}{2}\|{\bm y}-{\bm W}_2{\bm z}\|^2+\frac{\bar{\rho}}{2}\|{\bm z} -\sigma({\bm W}_1{\bm a}_0)\|^2.
\end{aligned}
\end{equation}
The  gradients  enjoy  the following form
\begin{align*}
\nabla_{{\bm W}_1} f&=\bar{\rho}\bm{a}_0^{\top}\textrm{Diag}[\sigma'({\bm W}_1{\bm a}_0)](\sigma({\bm W}_1{\bm a}_0)-{\bm z}),\\
 \nabla_{{\bm W}_2} f&={\bm z}^{\top}({\bm W}_2{\bm z}-{\bm y}),\\
 \nabla_{{\bm z}} f&={\bm W}_2^{\top}({\bm W}_2{\bm z}-{\bm y})+\bar{\rho}({\bm z} -\sigma({\bm W}_1{\bm a}_0)).
\end{align*}
 The gradient is then bounded by
$$\mathcal{O}\Big[(\|{\bm W}_2\|+\|{\bm z}\|+1)\cdot(\|{\bm W}_2{\bm z}-{\bm y}\|+1)+\bar{\rho}\|{\bm z} -\sigma({\bm W}_1{\bm a}_0)\|\Big].$$ The upper bound of the third-order gradient is also by bounded by these factors. For example, $\|\nabla_{{\bm z},{\bm z},{\bm W}_2} f\|=\mathcal{O}(\|{\bm W}_2\|)$, controlled by the upper bound of the gradient.
\subsection{The Nonconvex Accelerated SignSGD}
In the paper \cite{momentum2023}, the authors propose a single-loop nonconvex accelerated scheme as presented in Algorithm \ref{alg-sign}, also applying our settings due to the gradient normalization because it promises that $\|\bm{w}^{t+1}-\bm{w}^{t}\|$ can be sufficiently small to active the conditions in  Assumptions \ref{ass1} and \ref{ass2} if the learning rate $\gamma$ is set sufficiently small.
Algorithm \ref{alg-sign} indeed employs two kinds of momentums: step 1 uses Nesterov's momentum  \cite{nesterov1983method}, while step 2 uses Polyak's momentum \cite{polyak1964some,polyak1987introduction,ghadimi2016accelerated}.
These two momentums are recruited for different purposes; more specifically, Polyak's momentum is for convergence and Nesterov's momentum is for acceleration. Because step 3 iterates a gradient sign update rather than gradient descent, causing a bias of the stochastic gradient, which is fixed by Polyak's momentum.

One explanation for why momentum helps to guarantee convergence, in this case, is that it can be perceived as an alternative way to provide input from a ``large batch" because the deterministic sign gradient descent is convergent.
 This idea stems from the fact that momentum $\bm{m}^t$, through expanding the iterative equation $\bm{m}^t=\theta\bm{m}^{t-1}+(1-\theta)\nabla f(\bm{v}^t;\xi^t)$
, can be represented as $\bm{m}^t=\sum_{i=1}^t \alpha_i\nabla f(\bm{v}^i;\xi^i),$
 where $t$ denotes the iteration step and $(\alpha_i)_{1\leq i\leq t}$
 are coefficients given as $\alpha_i=\theta^{t-i}(1-\theta)$, where
 $\theta^{t-i}$ being the
$(t-i)$-th power of momentum parameter.
As $t$ gets larger, the sequence $(\bm{v}^i)_{i\geq 0}$
 tends to stabilize, and the coefficient $\alpha_i$
 will tend to 0 for small
$i$s. Therefore, one can find a large positive integer
 $B<t$ such that the approximation
\begin{equation}\label{batch-b}
\bm{m}^t\approx\sum_{i=1}^B \alpha_i\nabla f(\bm{v}^i;\xi^i)
\end{equation}
 is valid,  which shows that $\bm{m}^t$
 approximately contains the information for a
$B$-size batch. This guides momentum scheduling as it indicates that to find a large
$B$, we cannot make $\theta$
 too small which would lead  $(\alpha_i)_{1\leq i\leq t}$
 decay too quickly.

  \begin{algorithm}[H]
    \caption{Accelerated Sign Stochastic Gradient Descent (A-SignSGD)}\label{alg-sign}
	\begin{algorithmic}[1]
\REQUIRE   parameters $\gamma>0$,  $0\leq \theta<1$, $\zeta\geq 0$\\
\textbf{Initialization}: $\bm{w}^0=\bm{w}^1$, $\bm{m}^0=\bm{0}$\\
\textbf{for}~$t=1,2,\ldots$ \\
~~\textbf{step 1} $\bm{v}^t=\bm{w}^{t}+\zeta(\bm{w}^{t}-\bm{w}^{t-1})$\\
~~\textbf{step 2}: Sample the data  $\xi^t\sim\mathcal{D}$ and \\
~~~~~~~~~~~~~~$\bm{m}^t=\theta\bm{m}^{t-1}+(1-\theta)\nabla f(\bm{v}^t;\xi^t)$\\
~~\textbf{step 3}: $ \bm{w}^{t+1}= \bm{w}^{t}-\gamma \SI(\bm{m}^t)$ \\
\textbf{end for}\\
	\end{algorithmic}
  \end{algorithm}

\textbf{Assumptions on the noise:}
The analysis needs the bounded variance assumption for the stochastic gradient frequently used in this community \cite{bottou2018optimization}.
\begin{assumption}\label{ass3}
There exists a constant $\sigma>0$ such that the distribution $\mathcal{D}$ follows
$$
\sup_{\bm{w}\in\RR^d}\{\EE_{\xi\sim \mathcal{D}}\|\nabla f(\bm{w};\xi)-\nabla f(\bm{w})\|^2\}\leq \sigma^2.
$$
\end{assumption}

\section{ Convergence and Nonconvex Acceleration under Weak First- and Second-Order Gradient Lipschitz}
%
%
%

We present the convergence and nonconvex acceleration of A-SignSGD under weak smoothness assumptions.
    \begin{theorem}\label{th4}
1. Let $(\bm{w}^t)_{t\geq0}$ be generated by the A-SignSGD and Assumptions \ref{ass1}, \ref{ass3} and  \ref{ass4} hold. When $\zeta=0$,
for integer $T\geq \max\{\frac{(16L_2)^4d^2}{L_1^4},(\frac{L_2d}{L_1})^{\frac{4}{3}},(L_1r)^{-\frac{4}{3}}\}$, $1-\theta=\frac{1}{T^{1/2}}$,   $\gamma=\frac{1}{L_1T^{3/4}}$,    it holds that
 \begin{align*}
&\frac{1}{T}\sum_{t=1}^T\EE\|\nabla f(\bm{w}^{t})\|_1\leq \frac{4L_1(f(\bm{w}^{0})-\min f)}{T^{1/4}}+
\frac{2d}{T^{3/4}}\\
 &\qquad+\frac{8\sqrt{d}}{T^{1/4}} +\frac{8\sqrt{d}\|\nabla f(\bm{w}^0)\|}{\sqrt{T}}+\frac{16\sqrt{d}\sigma}{T^{1/4}}.
\end{align*}
2. Furthermore, let Assumption  \ref{ass2}  hold and  $\zeta=\frac{\theta}{1-\theta}$. For integer $T\geq\max\{ (\frac{dL_2}{C})^{\frac{7}{5}},
(rC)^{-\frac{7}{5}},
(RC)^{-\frac{7}{5}}, (\frac{8H_2}{C^2})^{7/2},128d^{21/4}\}$, $1-\theta=\frac{1}{T^{4/7}}$,  $\gamma=\frac{1}{CT^{5/7}}$,   it holds that
\begin{align*}
&\frac{1}{T}\sum_{t=1}^T\EE\|\nabla f(\bm{w}^{t})\|_1\\
&\quad\leq \frac{4\max\{\sqrt{H_1},\sqrt{H_2},L_1\}(f(\bm{w}^{0})-\min f)}{T^{2/7}}\\
&\qquad +\frac{4d^{3/2}}{T^{2/7}}+\frac{8\sigma}{T^{2/7}}+\frac{4d}{T^{5/7}}+\frac{8\sqrt{d}\|\nabla f(\bm{w}^0)\|}{T^{3/7}},
\end{align*}
where $C:=\max\{\sqrt{H_1},\sqrt{H_2},L_1\}$.
\end{theorem}

Theorem \ref{th4} consists of two sub-results: the convergence of A-SignSGD ($\zeta=0$) and the acceleration of A-SignSGD ($\zeta=\frac{\theta}{1-\theta}$). Unlike the results in \cite{momentum2023}, Theorem \ref{th4} introduces a significant novelty by removing both the first- and second-order smoothness assumptions.
It should be noted that the convergence rate of A-SignSGD is affected by the dimension of the space as each update only utilizes the sign information of the stochastic gradients.

For the convergence result where $\zeta=0$, the convergence rate is $\mathcal{O}\left(\frac{\sqrt{d}}{T^{1/4}}\right)$ for $\min_{1\leq t\leq T}\EE\|\nabla f(\bm{w}^t)\|_1$.
We can observe that $\|\cdot\|_1\leq \sqrt{d}\|\cdot\|$, and we know that the convergence rate for SGD is $\min_{1\leq t\leq T}\EE\|\nabla f(\bm{w}^t)\|=\mathcal{O}\left(\frac{1}{T^{1/4}}\right)$. Thus, by using the inequality $\|\cdot\|_1\leq \sqrt{d}\|\cdot\|$, we can derive
$
\min_{1\leq t\leq T}\EE\|\nabla f(\bm{w}^t)\|_1 \leq \sqrt{d}\min_{1\leq t\leq T}\EE\|\nabla f(\bm{w}^t)\|
=\mathcal{O}\left(\frac{\sqrt{d}}{T^{1/4}}\right).
$
Hence, we can conclude that the convergence rate for $\min_{1\leq t\leq T}\mathbb{E}\|\nabla f(\bm{w}^t)\|_1$ is also $\mathcal{O}\left(\frac{\sqrt{d}}{T^{1/4}}\right)$, which is the same as the convergence rate for sign-based method under weak first-order gradient Lipschitz.

Theorem \ref{th4} also demonstrates that the sign-based approach can achieve a faster convergence rate of $\mathcal{O}(\frac{d^{3/2}}{T^{2/7}})$ for $\min_{1\leq t\leq T}\EE\|\nabla f(\bm{w}^t)\|_1$ without relying on the first- and second-order Lipschitz smoothness assumptions. This finding enhances our understanding of the efficiency of sign-based algorithms.
It is worth mentioning that the convergence rate obtained from Theorem \ref{th4} matches the rates achieved in previous works \cite{tripuraneni2018stochastic,allen2018natasha,fang2019sharp,zhou2018finding,cutkosky2020momentum} for finding first-order stationary points under first- and second-order smoothness assumptions. However, Theorem \ref{th4} relaxes these smoothness assumptions substantially, making A-SignSGD applicable to a broader range of practical tasks.


Indeed, Theorem \ref{th4} provides the advantage of A-SignSGD compared to other nonconvex  accelerated algorithms when the second-order gradients are exploding, that is
\begin{equation}\label{boundg}
    \sup_{\bm{x}\in\RR^d}\|\nabla f(\bm{x})\|\leq g,~\sup_{\bm{x}\in\RR^d}\|\nabla^2 f(\bm{x})\|_{\textrm{op}}\leq G
\end{equation}
 with $g,G>0$.
 In this case, the weak first- and second-order smoothness also indicate
 the $L_1+L_2g$ first-order Lipschitz and $H_1+H_2G$ second-order Lipschitz constants, respectively. Take the proved result in \cite{tripuraneni2018stochastic} for example, in which the rate is
 \begin{align*}
 &\mathcal{O}\Big[\frac{d^{7/4}\sqrt{H_1+H_2G}}{\epsilon^{3.5}}+\frac{d^{3/2}(L_1+L_2g)}{(H_1+H_2G)\epsilon^3}\Big]\\
 &\qquad=\mathcal{O}\Big[\frac{d^{7/4}\sqrt{G}}{\epsilon^{3.5}}+\frac{d^{3/2}g}{G\epsilon^{3}}\Big]
 \end{align*}
 to reach the error $\min_{1\leq t\leq T}\mathbb{E}\|\nabla f(\bm{w}^t)\|_1\leq\epsilon$ when $\epsilon>0$ is small and $\min\{G,g\}\gg \max\{L_1,L_2,H_1,H_2,d^{7}\}$.
1) When $G$ is very large: While
 the complexity  for A-SignSGD is $\mathcal{O}(\frac{d^{\frac{21}{4}}}{\epsilon^{3.5}})$, which is much better than $\mathcal{O}(\frac{d^{7/4}\sqrt{G}}{\epsilon^{3.5}})$ if $G$ is very large.
 2) When $g$ is very large such that $g\gg\frac{d^{1/4}G^{3/2}}{\epsilon^{1/2}}$, the rate   is then $\mathcal{O}(\frac{d^{3/2}g}{G\epsilon^{3}})\gg\mathcal{O}(\frac{d^{7/4}\sqrt{G}}{\epsilon^{3.5}})\gg\mathcal{O}(\frac{d^{\frac{21}{4}}}{\epsilon^{3.5}})$, which also indicates the convergence advantage of  AN-SGD.

\medskip
\textbf{An understanding of the LION:} Below, we show that Theorem \ref{th4} also presents a theoretical explanation of the advantage of LION  \cite{chen2023symbolic}.
Consider the simplified  LION algorithm ($\beta=1$ in Algorithm 2 in \cite{chen2023symbolic}) updates
\begin{equation}\label{lion}
    \bm{x}^{t+1}= \bm{x}^{t}-\underbrace{\gamma\lambda\bm{x}^{t}}_{(\dag)}-\gamma \SI(\widetilde{\bm{m}}^t),
\end{equation}
where $\lambda\geq0$ and $\widetilde{\bm{m}}^t=\theta\widetilde{\bm{m}}^{t-1}+(1-\theta)\nabla f(\bm{x}^t;\xi^t)$. We turn back to the updating  of A-SignSGD,
\begin{equation}\label{lion-connect}
\begin{aligned}
    \bm{v}^{t+1}&= (1+\zeta)\bm{w}^{t+1} -\zeta\bm{w}^{t}\\
    &=\bm{w}^{t}-\gamma(1+\zeta)\SI(\bm{m}^t)\\
    &=\bm{v}^t-\underbrace{\gamma\zeta\SI(\bm{m}^{t-1})}_{(\ddag)}-\gamma(1+\zeta)\SI(\bm{m}^t).
    \end{aligned}
\end{equation}

We can see that LION is very similar to A-SignSGD except the term (\dag) in \eqref{lion} is replaced with $(\ddag)$ in \eqref{lion-connect}.
Note that when $\lambda=\zeta=0$, both schemes reduce to the SignSGD with momentum, whose convergence rate is as fast as $\mathcal{O}(\frac{1}{T^{1/4}})$ even without the first-order smoothness as presented in Theorem \ref{th4}.
When $\lambda>0$, without loss of generality, assume $\bm{x}^{0}=\textbf{0}$.
The scheme of the algorithm gives us
\begin{equation}\label{lion2}
    \bm{x}^{t}=-\gamma\sum_{j=0}^{t-1}(1- \gamma\lambda)^j\SI(\widetilde{\bm{m}}^{t-1-j}).
\end{equation}
As $t$ is large, the sequence $(\widetilde{\bm{m}}^i)_{i\geq 0}$
 tends to be stabile as $\widetilde{\bm{m}}^{t-1}$, based on which we assume $\SI(\widetilde{\bm{m}}^i)\approx \SI(\widetilde{\bm{m}}^{t-1})$ when $i$ is large.
 Furthermore, because the coefficient $(1- \gamma\lambda)^i$
is close to $0$ when $i$ is large,  there exists $M$ such that the approximation
\begin{align*}
\bm{x}^t&\approx\sum_{j=0}^{M-1} (1- \gamma\lambda)^j\SI(\widetilde{\bm{m}}^{t-1})\\
&=\frac{1-(1- \gamma\lambda)^{M}}{\gamma\lambda}\SI(\widetilde{\bm{m}}^{t-1}).
\end{align*}
In this case, the scheme of LION is then approximated by
\begin{equation}\label{lion2}
    \bm{x}^{t+1}\approx \bm{x}^{t}-(1-(1- \gamma\lambda)^{M})\SI(\widetilde{\bm{m}}^{t-1})-\gamma \SI(\widetilde{\bm{m}}^t),
\end{equation}
indicating that
 LION is very close to A-SignSGD except for the hyper-parameters in this case. Because  A-SignSGD is proved to achieve acceleration under weak first- and second-order Lipschitz smoothness, the approximation \eqref{lion2} then explains the possible advantage of LION to some extent.
\section{Extension to Nonconvex Distributed Settings}
This section contains the variant of the distributed normalized methods with the communication-efficient compressed operator.
There are numerous variants of SGD due to different settings of the specific problem, one of which requires compression before the gradient update.
Let  $\mathcal{C}:\RR^d\mapsto\RR^d$ be the compressed operator, i.e., there exists $\delta\in(0,1]$ such that
\begin{equation}\label{compress}
\|\CC(\bm{x})-\bm{x}\|^2\leq (1-\delta)\|\bm{x}\|^2,\forall \bm{x}\in\RR^d.
\end{equation}
A classical example is the top-$k$ sparsified operator \cite{alistarh2018convergence}, which only keeps the $k$  largest absolute elements and sets others to zeros.
It is easy to verify that the factor $\delta=\frac{s}{d}$ for a $d$-dimensional vector and $\delta=1$ indicates the identity operator, i.e., compression-free.

 We employ the fast compression communication (FCC) operator
in distributed training system \cite{huanglower}.
Given an integer $u\in\mathbb{Z}^+$, the FCC  operator with   $u$ rounds, $\textrm{FCC}_{u}:\RR^d\rightarrow\RR^d$ is defined as:
$$\textrm{FCC}_{u}(\bm{x}):=\sum_{k=0}^{u-1}\bm{c}^k,$$
where $\bm{v}^0=\bm{0}$, and $\bm{c}^k=\CC(\bm{x}-\bm{v}^k)$, $\bm{v}^{k+1}=\bm{v}^k+\bm{c}^k$ as $k\in[0,u-1]$. In \cite{huanglower}, the authors show that
\begin{equation}\label{comp}
    \EE\|\textrm{FCC}_{u}(\bm{x})-\bm{x}\|^2\leq(1-\delta)^u\|\bm{x}\|^2.
\end{equation}
This section considers the distributed algorithm for the following training task
 \begin{equation}\label{model-2}
    \min_{\bm{x}\in\RR^d} f(\bm{x})=\frac{1}{n}\sum_{i=1}^n \Big[f_i(\bm{x}):=\EE_{\xi\sim\mathcal{D}_i} f_i(\bm{x};\xi_i)\Big],
 \end{equation}
where $\xi\sim \mathcal{D}_i$. In the distributed setting, we need an extra widely used assumption.
\begin{assumption}\label{ass4}
There exists a constant $\bar{\sigma}>0$ such that the distribution $\mathcal{D}$ follows
$$
\sup_{\bm{x}\in\RR^d}\{\sum_{i=1}^n\|\nabla f_i(\bm{x})-\nabla f(\bm{x})\|^2/n\}\leq \bar{\sigma}^2.
$$
\end{assumption}
  \begin{algorithm}[H]
    \caption{    Compressed Accelerated Sign Stochastic Gradient Descent (CA-SignSGD)}\label{alg-dis}
	\begin{algorithmic}[1]
\REQUIRE   parameters $\gamma>0$,  $0\leq \theta<1$, $\zeta\geq 0$\\
\textbf{Initialization}: $\bm{w}^0=\bm{w}^1$, $\bm{m}^0=\bm{0}$\\
\textbf{for}~$t=1,2,\ldots$ \\
~~\textbf{step 1}: Worker $i\in[n]$ receives $\bm{w}^t$ from the PS to get\\
~~~~~~~~~~~~~~$\bm{v}^t=\bm{w}^{t}+\zeta(\bm{w}^{t}-\bm{w}^{t-1})$\\
~~\textbf{step 2}: Worker $i$ samples $\xi_i^t\sim \mathcal{D}_i$ to calculate\\
~~~~~~~~~~~~~~$\bm{g}^t(i)=\nabla f_i(\bm{w}^t;\xi_i^t)$  \\
~~\textbf{step 3}: Worker $i$ sends $\bar{\bm{g}}^t(i)=\textrm{FCC}_{u}(\bm{g}^t(i))$ to PS\\

~~\textbf{step 4}: PS runs $\bm{m}^t=\theta\bm{m}^{t-1}+(1-\theta)\sum_{i=1}^n \bar{\bm{g}}^t(i)/n$\\
~~\textbf{step 5}: PS runs $ \bm{w}^{t+1}= \bm{w}^{t}-\gamma \SI(\bm{m}^t)$ \\
\textbf{end for}\\
	\end{algorithmic}
  \end{algorithm}
  \begin{theorem}\label{th3}

1. Let $(\bm{w}^t)_{t\geq0}$ be generated by the CA-SignSGD and Assumptions \ref{ass1}, \ref{ass3} and  \ref{ass4} hold. When $\zeta=0$,
for integer $T\geq \max\{\frac{(16dL_2)^4}{L_1^4},(\frac{L_2d}{L_1})^{\frac{4}{3}},(L_1r)^{-\frac{4}{3}}\}$, $1-\theta=\frac{\sqrt{n}}{T^{1/2}}$,   $\gamma=\frac{n^{1/4}}{L_1T^{3/4}}$, and $u=\frac{2\ln(64\sqrt{2d}nT^{1/4})}{\ln(\frac{1}{1-\delta})}$  \footnote{We use the convention $\frac{1}{0}:=+\infty$.},  it holds that
 \begin{align*}
&\frac{1}{T}\sum_{t=1}^T\EE\|\nabla f(\bm{w}^{t})\|\\
&\quad\leq \frac{4L_1(f(\bm{w}^{0})-\min f)}{n^{1/4}T^{1/4}}+\frac{16d}{n^{1/4}T^{1/4}} +\frac{8\sqrt{d}\|\nabla f(\bm{w}^0)\|}{\sqrt{nT}}\\
&\quad\qquad+\frac{16\sqrt{d}\sigma}{n^{1/4}T^{1/4}}+\frac{\sqrt{\sigma^2+2\bar{\sigma}^2}}{8nT^{1/4}}.
\end{align*}
2. Furthermore, let Assumption  \ref{ass2}  hold  and  $\zeta=\frac{\theta}{1-\theta}$. For integer $T\geq\max\{ (\frac{dL_2}{C})^7,
(rC)^{-\frac{7}{5}},
(RC)^{-\frac{7}{5}}, (\frac{8d^{3/2}H_2}{C^2})^{7/2},n^4\}$, $1-\theta=\frac{n^{1/2}}{T^{4/7}}$,  $\gamma=\frac{n^{1/4}}{CT^{5/7}}$, and $u=\frac{2\ln[(64\sqrt{2}+64\sqrt{2d})nT]}{\ln\frac{1}{1-\delta}}$,  it holds that
 \begin{align*}
&\frac{1}{T}\sum_{t=1}^T\EE\|\nabla f(\bm{w}^{t})\|\\
&\quad\leq \frac{4C(f(\bm{w}^{0})-\min f)}{n^{2/7}T^{2/7}}+\frac{3d}{n^{2/7}T^{4/7}}+\frac{4\sqrt{d}}{n^{4/7}T^{2/7}}\\
 &\quad\qquad+ \frac{16\sqrt{d}\|\nabla f(\bm{w}^0)\|}{n^{2/7}T^{3/7}}+\frac{8\sqrt{d}\sigma}{n^{2/7}T^{2/7}}+\frac{\sqrt{d}\sqrt{\sigma^2+2\bar{\sigma}^2}}{8nT^{2/7}},
\end{align*}
where constant $C$ follows the same definition in Theorem \ref{th4}.
\end{theorem}
Note that the proved results also contain the non-compressed case when $\delta=1$, where the round number reduces to $u=0$.
Theorem \ref{th3} presents the convergence of two sub-algorithms, i.e., the distributed SignSGD with FCC ($\zeta=0$) and its acceleration variant ($\zeta=\frac{\theta}{1-\theta}$) without first- and second-order Lipschitz smoothness. For both cases,  the rate can be improved as $n$ increases, unsurprisingly.
As $n$ and $T$ tend to infinity, distributed SignSGD with FCC achieves a convergence rate of $\mathcal{O}(\frac{\sqrt{d}(1+\sigma)}{n^{1/4}T^{1/4}})$ for $\min_{1\leq t\leq T}\EE\|\nabla f(\bm{w}^t)\|_1$, which is comparable to the rate obtained by previous distributed SGD with FCC discussed in \cite{huanglower}. However, it is important to note that in our analysis, we relied on a much weaker assumption of Lipschitz smoothness.
In Theorem \ref{th3}, the round number $u$ in the FCC is set as $\mathcal{O}(\frac{\ln T}{\delta})$, which keeps the same order in  \cite{huanglower}.

On the other hand, Theorem \ref{th3} also indicates the advantage of distributed SignSGD with FCC when the gradients are exploding. Assume that  the bounded assumption, i.e., $\sup_{\bm{x}\in\RR^d}\|\nabla f(\bm{x})\|\leq g$ giving the  $L_1+L_2g$ first-order Lipschitz smoothness when  Assumption  \ref{ass1} holds. The rate of distributed SGD with FCC is then $\mathcal{O}(\frac{\sqrt{d}(1+\sigma)\sqrt{L_1+L_2g}}{n^{1/4}T^{1/4}})=\mathcal{O}(\frac{\sqrt{d}(1+\sigma)\sqrt{g}}{n^{1/4}T^{1/4}})$ for $\min_{1\leq t\leq T}\EE\|\nabla f(\bm{w}^t)\|_1$, which is much worse than $\mathcal{O}(\frac{\sqrt{d}(1+\sigma)}{n^{1/4}T^{1/4}})$ as $g$ is very large.
Denote $\mathfrak{c}$ as the cost of once communication for the compression operator $\mathcal{C}$.
In this case, the total communication cost of distributed SGD with FCC is $\widetilde{\mathcal{O}}(\frac{(1+\sigma)^4g^2d^2\mathfrak{c}}{\delta\epsilon^4})$,  also underperforming than the sign  version whose communication cost is $\widetilde{\mathcal{O}}(\frac{(1+\sigma)^4d^2\mathfrak{c}}{\delta\epsilon^4})$ as $g$ is large.

While for $\zeta=\frac{\theta}{1-\theta}$, we proved the nonconvex acceleration when the weak second-order smoothness further holds.
As $n$ and $T$ are large, the improved rate can be as fast as $\mathcal{O}(\frac{\sqrt{d}(1+\sigma)}{n^{2/7}T^{2/7}})$,  much faster than that of distributed (sign) SGD with FCC. The communication cost  of accelerated nonconvex algorithm is $\widetilde{\mathcal{O}}(\frac{(1+\sigma)^{7/2}d^{7/4}\mathfrak{c}}{\delta\epsilon^{7/2}})$, achieving significant reduction.
%
%
%
%

\section{Numerics}

In this section, we aim to provide numerical evidence to validate our theoretical findings. However, in real-world applications, training often involves additional modifications such as weight decay. Consequently, the training performed may no longer strictly adhere to the original scheme.
To address this, we conduct two types of experiments. Firstly, we demonstrate the convergence and superiority of A-SignSGD compared to other training methods when all models are trained using the original scheme, i.e. without weight decay. Secondly, we extend our evaluation by comparing our algorithm to previous baselines in different settings with weight decay.

\textbf{Dataset, Model, and Baseline.} We use ResNet18 \cite{he2016deep} with CIFAR100 \cite{krizhevsky2009learning} as the benchmarks of our method.
The models on  CIFAR100 are trained with 128 batch size.
In this paper, we compare A-SignSGD, with several baseline algorithms, including SGDm \cite{sutskever2013importance}, Adam \cite{DBLP:journals/corr/KingmaB14}, SignSGDM \cite{pmlr-v80-bernstein18a}, and Lion \cite{chen2023symbolic}.

\textbf{Hyperparameters.} For the purpose of our evaluation, we implement all of these algorithms using their original schemes, with the recommended hyperparameters of momentums as provided in the respective papers\footnote{The recommended momentums for SGDm and Adam can be found in the PyTorch documentation. The recommended momentums for LION can be obtained from the code link provided in the paper by \cite{chen2023symbolic}. For SignSGDM, the recommended monemtums can be found in Figure A.4 of the paper by \cite{pmlr-v80-bernstein18a}.}. In the experiment, for A-SignSGD, we set the values of $\theta$ and $\zeta$ to be 0.9 and 0.99, respectively.
 The learning rate   is   multiplied by 0.2 after every 30 epochs.

All figures in the experiments are plotted by MATLAB 2018a.
\subsection{Validation of Theory}
\begin{figure*}
      \centering
      \subcaptionbox{Test Accuracy}
      {\includegraphics[scale=0.4]{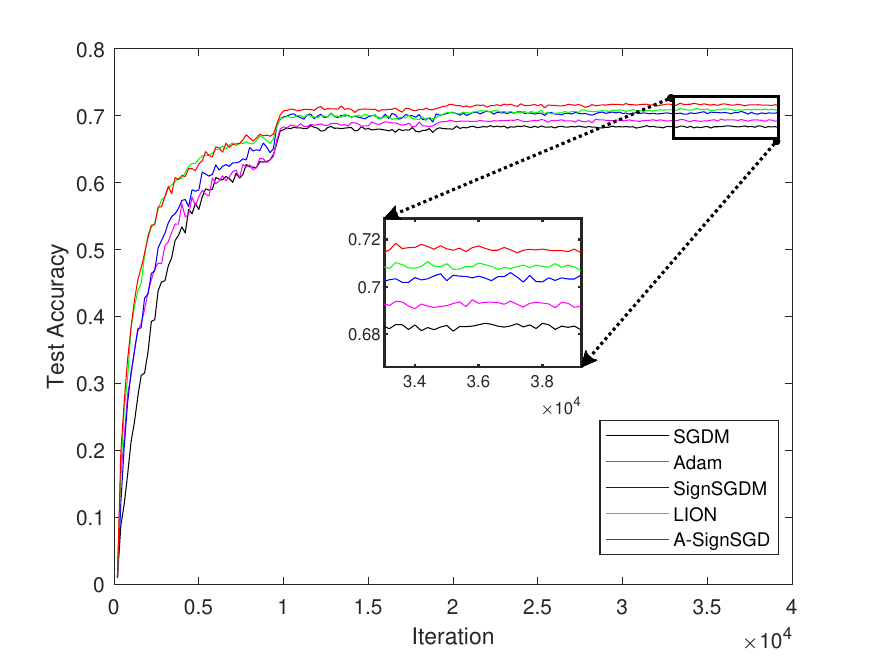}}
      \hfill
      \subcaptionbox{Training Loss}
        {\includegraphics[scale=0.4]{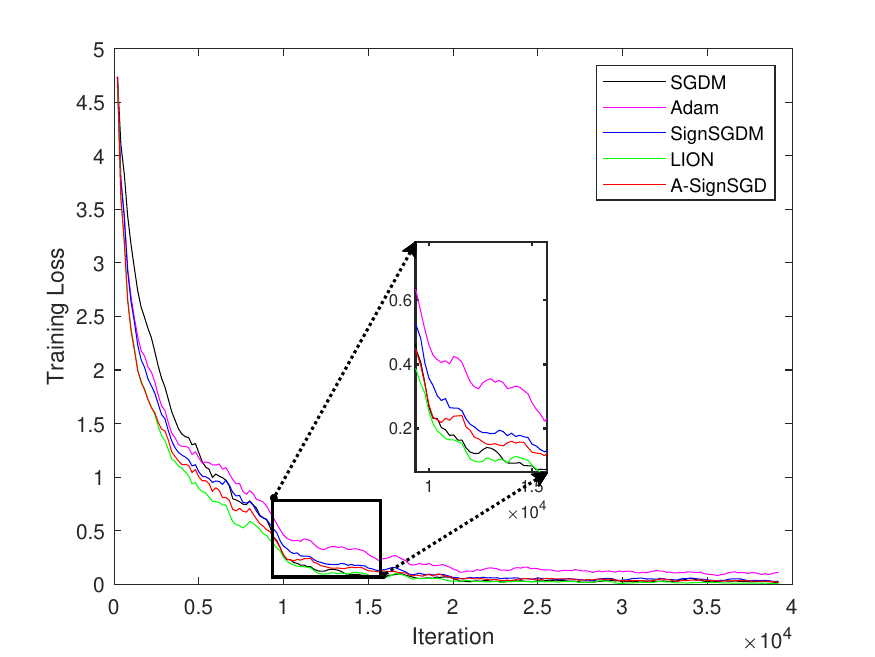}}
        \hfill
      \subcaptionbox{Test Loss}
        {\includegraphics[scale=0.4]{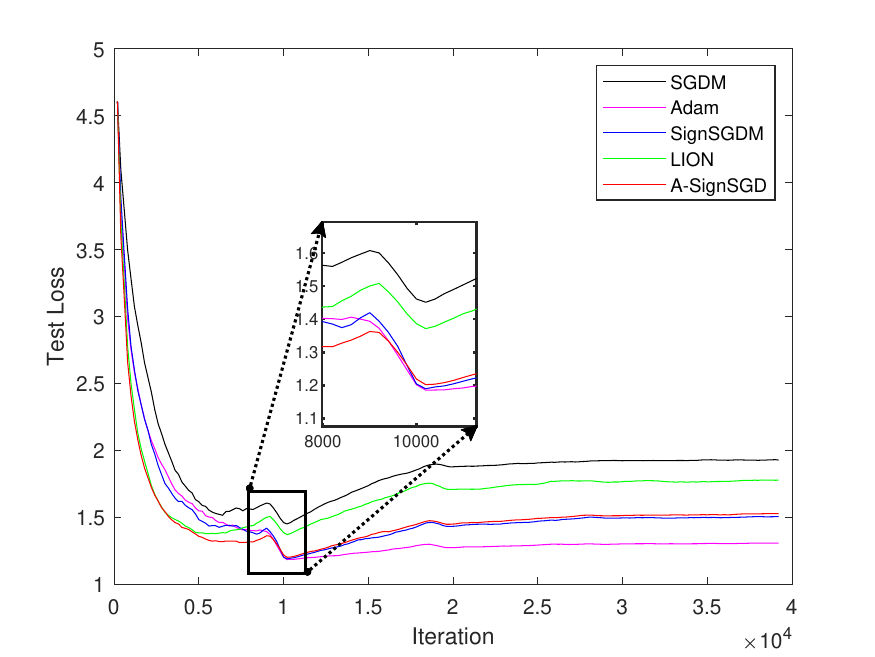}}
            \caption{Performance profiles of A-SignSGD, SGDM, Adam, SignSGDM, and Lion with ResNet-18 model on CIFAR 100 dataset. All algorithms were implemented without weight decay on single machine.}
        \label{fig:single-cifar}
\end{figure*}

We tune the initial learning rates from $\{0.1, 0.01, 0.001, 0.0001\}$ for all algorithms.
 The
optimal initial learning rates for SGDMW is 0.1, and AdamW is 0.001, others algorithms are 0.0001.
The models were trained for 100 epochs using different algorithms, and the results are shown in Fig. \ref{fig:single-cifar}.
It was observed that A-SignSGD outperformed the other baselines in terms of test accuracy. Moreover, A-SignSGD showed comparable performance to LION, SGDM and SignSGDM, and it was faster than  Adam in terms of training loss. It is important to highlight that our implementation does not include weight decay to align with the original schemes of all algoroithms.
Additionally, it is worth noting that A-SignSGD is theoretically supported in cases of weak smoothness.

\begin{figure*}
      \centering
      \subcaptionbox{Test Accuracy}
      {\includegraphics[scale=0.4]{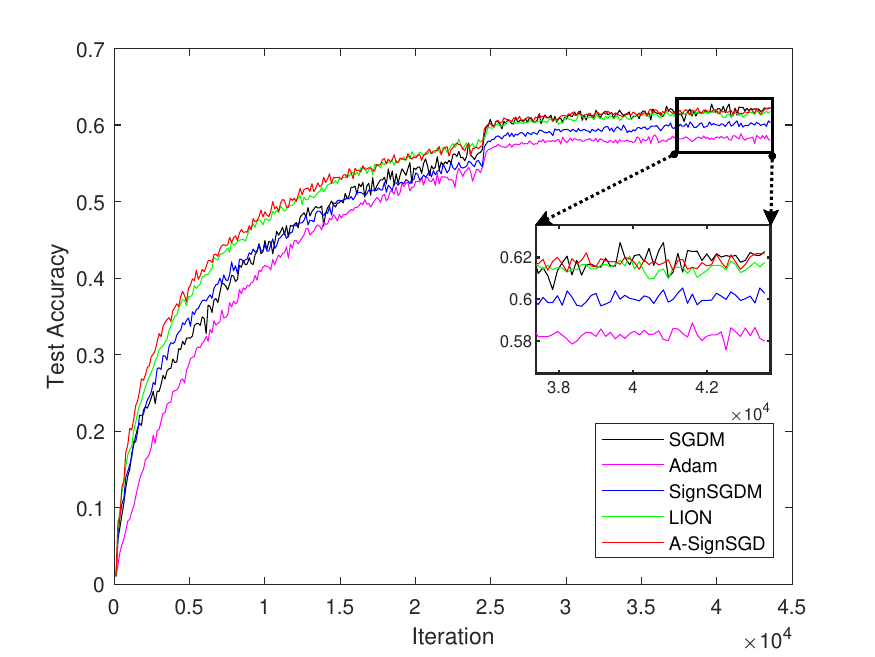}}
      \hfill
      \subcaptionbox{Training Loss}
        {\includegraphics[scale=0.4]{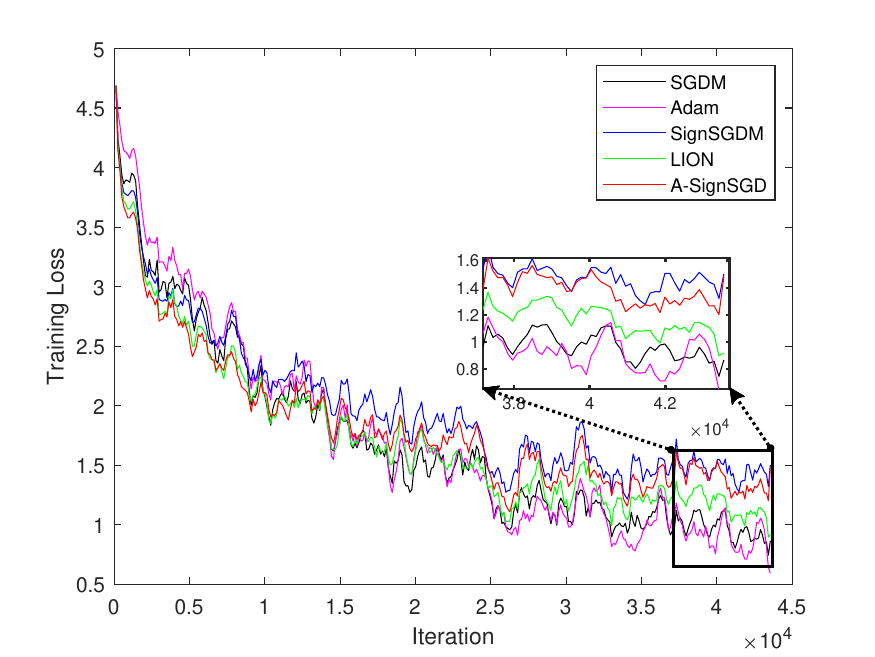}}
        \hfill
      \subcaptionbox{Test Loss}
        {\includegraphics[scale=0.4]{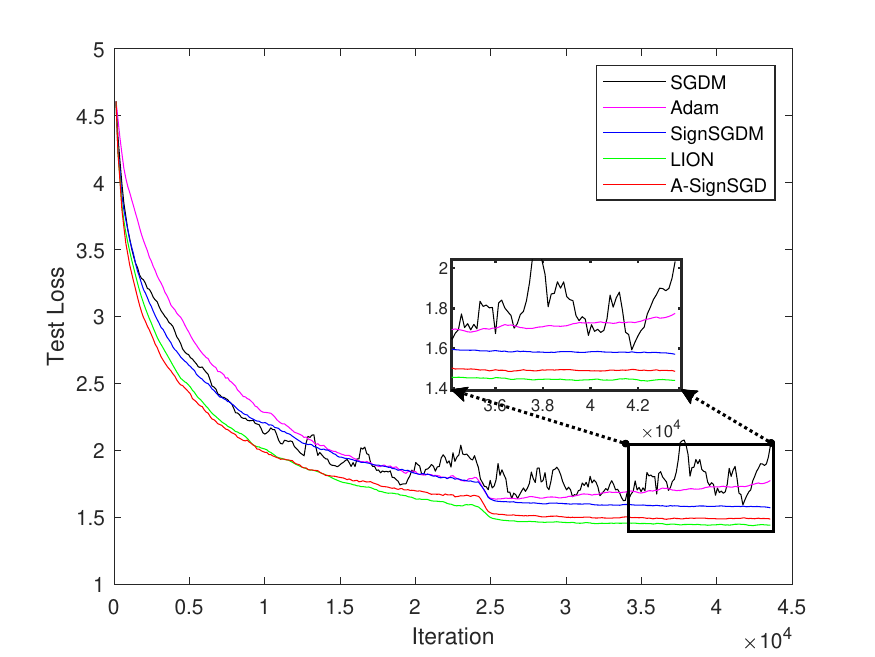}}
            \caption{Performance profiles of A-SignSGD, SGDM, Adam, SignSGDM, and Lion with ResNet-18 model on CIFAR 100 dataset. All algorithms were implemented without weight decay on distributed scenario with compression.}
        \label{fig:dis-cifar}
\end{figure*}

In the distributed scenario, we explore the performance of different algorithms using a Top-5\% compressor and setting $u=2$ in FCC. Our network simulation consists of 4 nodes, where each node has independently and identically distributed data from the training dataset. We also conduct tuning of the initial learning rates for all algorithms from a set of values: $\{0.1, 0.01, 0.001, 0.0001\}$.
The optimal initial learning rate for the SGDM algorithm is determined to be 0.1, while for the Adam algorithm it is found to be 0.01. For the other algorithms, the optimal initial learning rate is set to 0.0001.
Figure \ref{fig:dis-cifar} displays the outcomes of the distributed scenario without weight decay. It can be observed that our proposed algorithm shows comparable performance to LION and SGDM, while demonstrating better test accuracy compared to SignSGD and Adam,  when employing the FCC with a Top-5\% compressor.

\subsection{Comparison with Weight Decays}
\begin{figure*}
      \centering
      \subcaptionbox{Test Accuracy}
      {\includegraphics[scale=0.4]{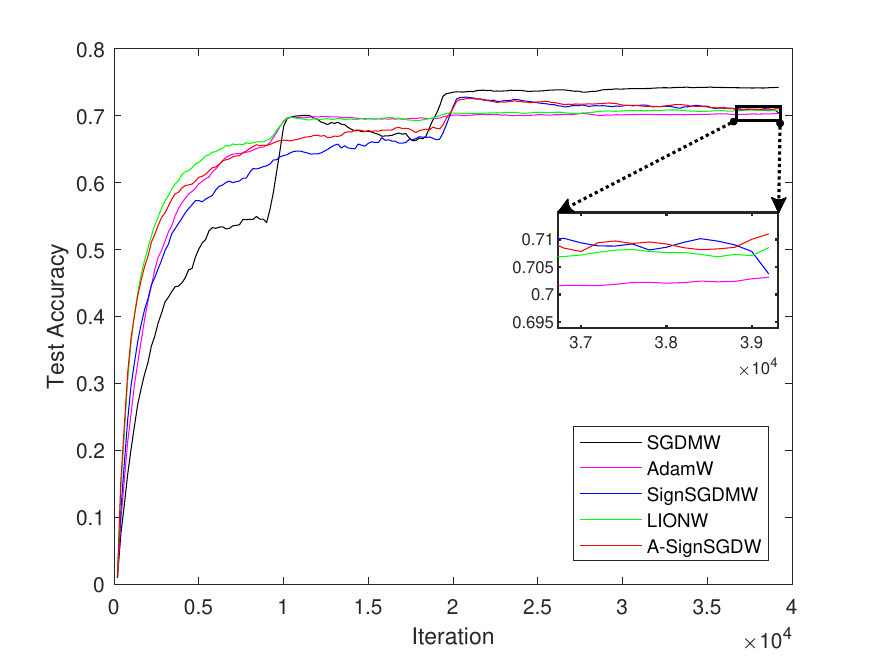}}
      \hfill
      \subcaptionbox{Training Loss}
        {\includegraphics[scale=0.4]{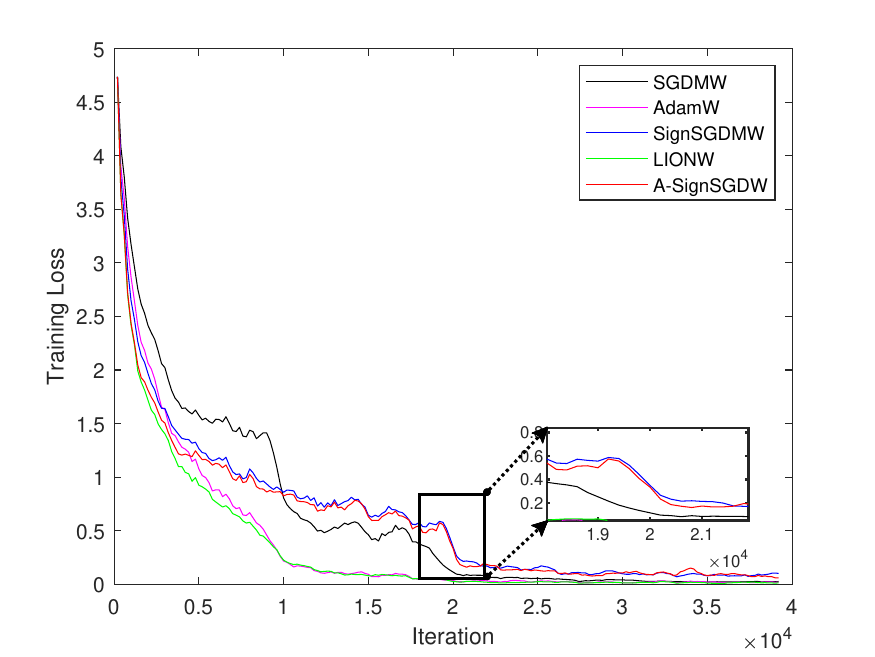}}
        \hfill
      \subcaptionbox{Test Loss}
        {\includegraphics[scale=0.4]{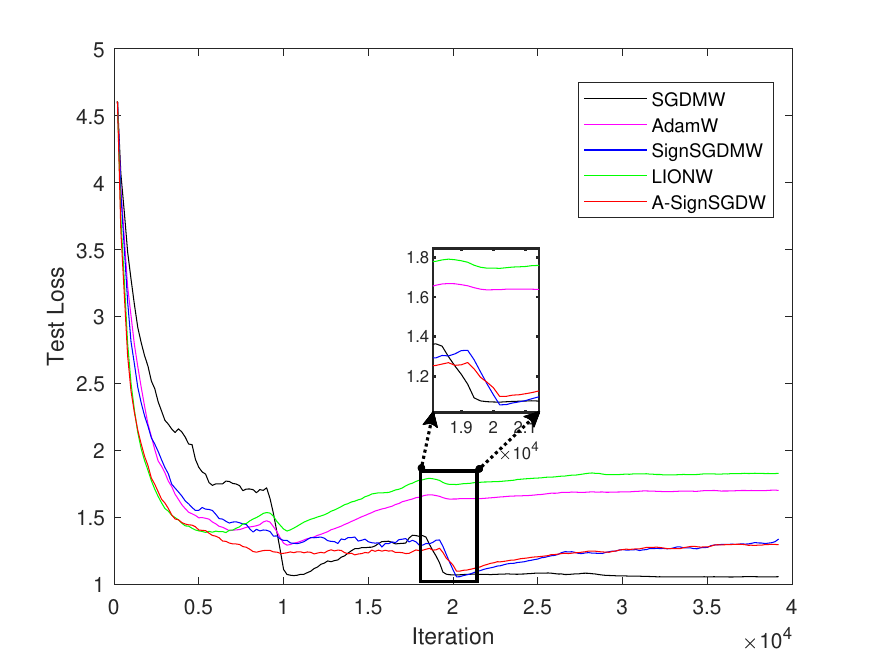}}
            \caption{Performance profiles of A-SignSGDW, SGDMW, AdamW, SignSGDMW, and LionW with ResNet-18 model on CIFAR 100 dataset. All algorithms were implemented with weight decay on single machine.}
        \label{fig:single-cifar-real}
\end{figure*}

This subsection considers the training with weight decay as 5.0e-4 for all algorithms in the  training. We conduct tuning of the initial learning rates for all algorithms from a set of values: $\{0.1, 0.01, 0.001, 0.0001\}$. After performing the necessary evaluations, we determine the optimal initial learning rates for SGDMW and AdamW to be 0.1 and 0.001, respectively. For the remaining algorithms, the optimal initial learning rate is set to 0.0001.
In the comparison of the single-machine case, the results are presented in Figure \ref{fig:single-cifar-real}. It is evident that SGDM exhibits the best performance when weight decay is applied. In this scenario, A-SignSGDW demonstrates superior performance compared to LIONW and AdamW. It also shows slightly better performance than SignSGDW in the final few training iterations.

\begin{figure*}
      \centering
      \subcaptionbox{Test Accuracy}
      {\includegraphics[scale=0.4]{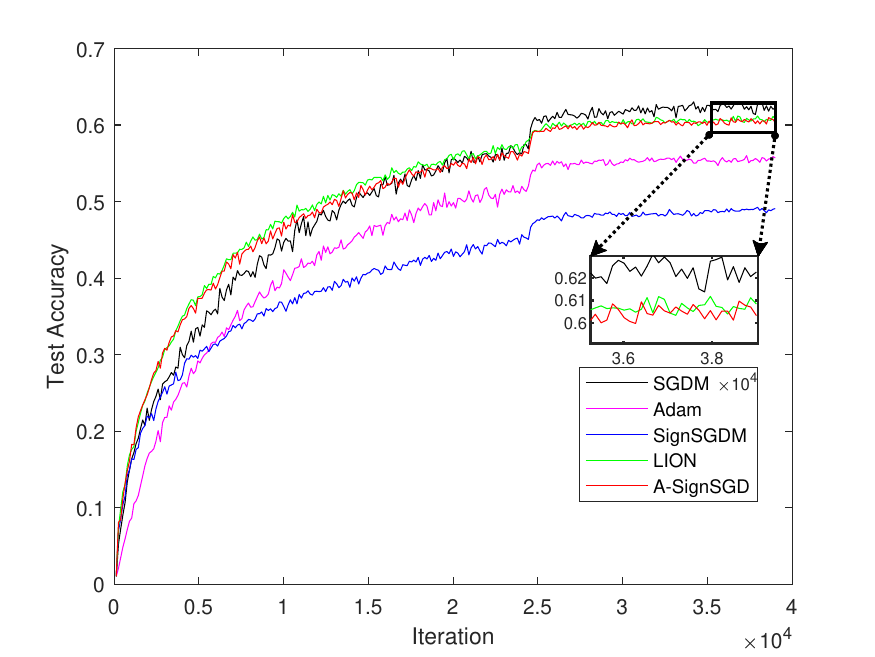}}
      \hfill
      \subcaptionbox{Training Loss}
        {\includegraphics[scale=0.4]{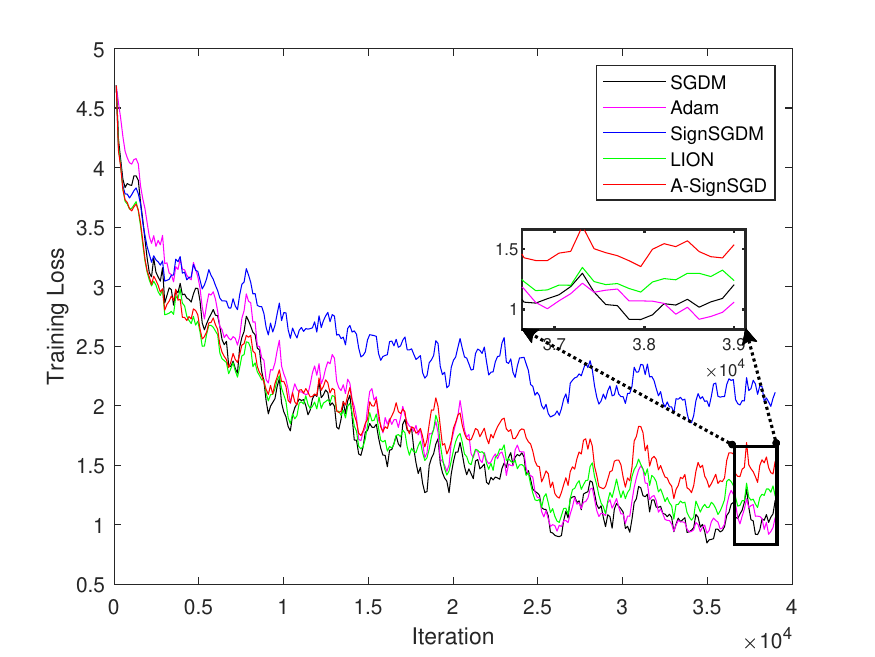}}
        \hfill
      \subcaptionbox{Test Loss}
        {\includegraphics[scale=0.4]{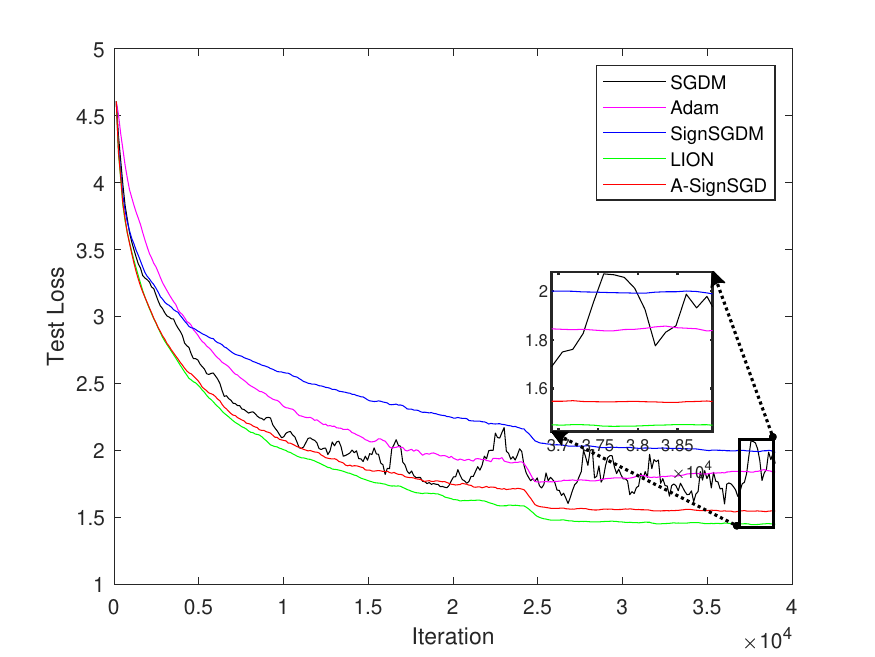}}
            \caption{Performance profiles of A-SignSGDW, SGDMW, AdamW, SignSGDMW, and LionW with ResNet-18 model on CIFAR 100 dataset. All algorithms were implemented with  weight decay on distributed scenario with compression.}
        \label{fig:dis-cifar-real}
\end{figure*}

In the distributed setting with decays, we also perform tuning of the initial learning rates. Specifically, we find that the optimal initial learning rates for SGDMW and AdamW are 0.1 and 0.01, respectively. For the remaining algorithms, we set the optimal initial learning rate to 0.0001.
In the distributed scenarios with compressions, the results can be found in Figure \ref{fig:dis-cifar-real}. Similar to the single-machine case, SGDMW exhibits the best performance. A-SignSGDW is compared to LIONW and outperforms the other algorithms.

\section{Concluding Remarks}
Our results can be further extended to a general nonconvex accelerated framework. We consider a general operator $\Tc:\RR^d\rightarrow\RR^d$ satisfying a condition as follows.

\medskip
\noindent\textbf{Condition 1.}
\textit{For any $\bm{x}\in\RR^d$, it holds
$$\langle\bm{x},\Tc(\bm{x})\rangle\geq l\|\bm{x}\|_{\diamond},~\|\Tc(\bm{x})\|\leq U,$$
where $l,U>0$, and $\|\cdot\|_{\diamond}$ is some norm of $\RR^d$.}
\medskip

Notably, the sign operator also satisfies Condition 1. With that in mind, the following proposition holds under Condition 1.
 \begin{proposition}\label{pro-cond}
Let $\bm{w}^{\dag},\bm{m}\in\RR^d$ be arbitrary vectors,  and
\begin{equation}\label{temp-general}
\bm{w}^{\ddag}= \bm{w}^{\dag}-\gamma \Tc(\bm{m}),
\end{equation}
and $\bm{\epsilon}:=\bm{m}-\nabla f(\bm{w}^{\dag})$. If Assumption \ref{ass1} holds, as $0<\gamma\leq\max\{\frac{al}{U^2L_2},r\}$, we have
\begin{align*}
f(\bm{w}^{\ddag})-f(\bm{w}^{\dag})&\leq -\frac{\gamma l}{2}\|\nabla f(\bm{w}^{\dag})\|_{\diamond}\\
&+(U+b)\gamma\|\bm{m}-\nabla f(\bm{w}^{\dag})\|+\frac{L_1U^2}{2}\gamma^2.
\end{align*}
\end{proposition}
We can present the accelerated general gradient normalization algorithm based on this result.
  \begin{algorithm}[H]
    \caption{  General    Sign SGD  (G-SignSGD)}\label{alg-GN}
	\begin{algorithmic}[1]
\REQUIRE   parameters $\gamma>0$,  $0\leq \theta<1$, $\zeta\geq 0$, operator $\Tc$\\
\textbf{Initialization}: $\bm{w}^0$, $\bm{m}^0=\bm{0}$\\
\textbf{for}~$t=1,2,\ldots$ \\
~~\textbf{step 1$\sim$2} same as A-SignSGD\\
~~\textbf{step 3}: $ \bm{w}^{t+1}= \bm{w}^{t}-\gamma \Tc(\bm{m}^t)$ \\
\textbf{end for}\\
	\end{algorithmic}
  \end{algorithm}

The convergence and acceleration of the G-SignSGD are then presented as follows.
 \begin{proposition}\label{pro-conver}
1. Let $(\bm{w}^t)_{t\geq0}$ be generated by the G-SignSGD and Assumptions \ref{ass1}, \ref{ass3} and  Condition 1 hold. When $\zeta=0$,
for sufficiently large $T$,   $\gamma=\frac{1}{L_1T^{3/4}}$,    it holds that
$\frac{1}{T}\sum_{t=1}^T\EE\|\nabla f(\bm{w}^{t})\|_{\diamond}\leq \mathcal{O}(\frac{1+\sigma}{T^{1/4}})$.
2. Furthermore, let Assumption  \ref{ass2}  hold and  $\zeta=\frac{\theta}{1-\theta}$, for sufficiently large $T$, $1-\theta=\frac{1}{T^{4/7}}$,  $\gamma=\frac{1}{CT^{5/7}}$,   it holds that
$
\frac{1}{T}\sum_{t=1}^T\EE\|\nabla f(\bm{w}^{t})\|_{\diamond}= \mathcal{O}(\frac{1+\sigma}{T^{2/7}}),
$
where constant $C$ follows the same definition in Theorem \ref{th4}.
\end{proposition}

%
%
%
%
%

\newpage
\onecolumn
\begin{center}
{\Large \bf Proofs for}
\end{center}
\vspace{-0.4cm}
\begin{center}
\large \bf \textit{Rethinking SIGN Training: Provable Nonconvex Acceleration without   First- and Second-Order   Gradient Lipschitz}

\end{center}
\appendices
\section{Technical Lemmas}
\begin{lemma}\label{lemma0}
When Assumption \ref{ass1} holds, for any $\bm{x},\bm{y}\in\RR^d$ such that $\|\bm{x}-\bm{y}\|\leq r$, we have
\begin{align*}
f(\bm{y})-f(\bm{x})\leq \langle\nabla f(\bm{x}),\bm{y}-\bm{x}\rangle+\frac{L_1+L_2\|\nabla f(\bm{x})\|}{2}\|\bm{y}-\bm{x}\|^2.
\end{align*}
When Assumption \ref{ass2} holds, for any $\bm{x},\bm{y}\in\RR^d$ such that $\|\bm{x}-\bm{y}\|\leq R$, we have
\begin{align*}
\|\nabla f(\bm{y})-\nabla f(\bm{x})-[\nabla^2 f(\bm{x})](\bm{y}-\bm{x})\|\leq \frac{H_1+H_2\|\nabla f(\bm{x})\|}{2}\|\bm{y}-\bm{x}\|^2.
\end{align*}
\end{lemma}

\begin{lemma}\label{lemma1-sign}
Let $\bm{w}^{\dag},\bm{m}\in\RR^d$ be arbitrary vectors,  and
\begin{equation}\label{temp-sch-sign}
\bm{w}^{\ddag}= \bm{w}^{\dag}-\gamma \SI(\bm{m}),
\end{equation}
and $\bm{\epsilon}:=\bm{m}-\nabla f(\bm{w}^{\dag})$. If Assumption \ref{ass1} holds, as $0<\gamma\leq\max\{\frac{1}{L_2d},r\}$, we have
\begin{align*}
f(\bm{w}^{\ddag})-f(\bm{w}^{\dag})\leq -\frac{\gamma}{2}\|\nabla f(\bm{w}^{\dag})\|_1+2\gamma\sqrt{d}\|\bm{m}-\nabla f(\bm{w}^{\dag})\|+\frac{L_1}{2}\gamma^2d.
\end{align*}
\end{lemma}

\begin{lemma}\label{lemma2}
Let $(\bm{v}^t)_{t\geq0}$ be generated by the CA-SignSGD, and Assumption \ref{ass3} hold, we have
$$ \EE_{\CC,\chi^t}\Big\|\sum_{i=1}^n \hat{\bm{g}}^t(i)/n-\nabla f(\bm{v}^t)\Big\|^2\leq (2\sigma^2+4\bar{\sigma}^2)(1-\delta)^u+\frac{2\sigma^2}{n}+4(1-\delta)^u\|\nabla f(\bm{v}^t)\|^2.$$
\end{lemma}

\begin{lemma}\label{lemma3}
Let $(\bm{v}^t)_{t\geq0}$ be generated by the CA-SignSGD, and Assumption \ref{ass3} hold, as $k\neq j$, we have
\begin{align*}
 &\EE_{\CC,\chi^t}\Big\langle\sum_{i=1}^n\bar{\bm{g}}^k(i)/n-\nabla f(\bm{v}^k),\sum_{i=1}^n\bar{\bm{g}}^j(i)/n-\nabla f(\bm{v}^j)\Big\rangle\\
 &\qquad\leq (2\sigma^2+4\bar{\sigma}^2)(1-\delta)^u+2(1-\delta)^u\|\nabla f(\bm{v}^k)\|^2+2(1-\delta)^u\|\nabla f(\bm{v}^j)\|^2.
\end{align*}
\end{lemma}

\section{Proof of Theorem \ref{th4}}
\textbf{Case 1: $\zeta=0$}.
We employ the short hand notation
\begin{equation}\label{th4-reuse-n}
 \bm{\epsilon}^t:=\bm{m}^t-\nabla f(\bm{w}^t), \bm{\delta}^t:=\bm{g}^t-\nabla f(\bm{w}^t),\bm{s}^t:=\nabla f(\bm{w}^{t-1})-\nabla f(\bm{w}^{t}).
\end{equation}
The scheme of A-SignSGD yields the following equation
\begin{align*}
\bm{m}^{t}=\theta\bm{m}^{t-1}+(1-\theta)\bm{g}^t=\theta\Big[\bm{\epsilon}^{t-1}+\nabla f(\bm{w}^{t-1})\Big]+(1-\theta)\Big[\bm{\delta}^t+\nabla f(\bm{w}^t)\Big],
\end{align*}
based on which we get
\begin{align*}
\bm{\epsilon}^t=\bm{m}^{t}-\nabla f(\bm{w}^t)=\theta\bm{\epsilon}^{t-1}+\theta\bm{s}^t+(1-\theta)\bm{\delta}^t.
\end{align*}
With the use of mathematical induction, we can further derive
\begin{align}\label{ada-reuse-th4}
  \bm{\epsilon}^t= \theta^t\bm{\epsilon}^{0}+\sum_{k=1}^{t}\theta^{t-k}\bm{s}^k+(1-\theta)\sum_{k=1}^{t}\theta^{t-k}\bm{\delta}^k.
\end{align}
By taking the norms of both sides of inequality \eqref{ada-reuse-th4},
\begin{align*}
   \EE\|\bm{\epsilon}^t\|\leq \sum_{k=1}^{t}\theta^{t-k}\EE\|\bm{s}^k\|+(1-\theta)\EE\Big\|\sum_{k=1}^{t}\theta^{t-k}\bm{\delta}^k\Big\|+\theta^t\|\bm{\epsilon}^{0}\|.
\end{align*}
As $\gamma\leq r$, the weak smooth property of function $f$ gives us
\begin{align*}
\EE\|\bm{s}^k\|\leq \EE\Big[(L_1+L_2\|\nabla f(\bm{w}^{k-1})\|)\|\bm{w}^{k}-\bm{w}^{k-1}\|\Big]=L_1\gamma+L_2\gamma\EE\|\nabla f(\bm{w}^{k-1})\|.
\end{align*}
Turning back to \eqref{ada-reuse-th4}, we derive
\begin{align}\label{wucha-th4}
\begin{aligned}
  \sum_{t=1}^{T}\EE\|\bm{\epsilon}^t\|/T
 &\leq\frac{ L_1\gamma}{1-\theta} +\frac{\|\nabla f(\bm{w}^0)\|}{(1-\theta)T}+\frac{L_2\gamma}{(1-\theta)}\frac{\sum_{t=1}^{T}\EE\|\nabla f(\bm{w}^{t})\|}{T}+2\sqrt{1-\theta}\sigma.
 \end{aligned}
\end{align}
Using Lemma \ref{lemma1-sign} with  $\bm{w}^{\dag}\rightarrow \bm{w}^{t}$ and $\bm{m}\rightarrow \bm{m}^{t}$ and taking taking expectations,
\begin{align*}
\EE (f(\bm{w}^{t+1}))-\EE f(\bm{w}^{t})\leq -\gamma/2 \EE\|\nabla f(\bm{w}^{t})\|_1+2\gamma\sqrt{d}\EE\|\bm{\epsilon}^t\|+\frac{L_1}{2}\gamma^2d.
\end{align*}
Summing the recursion from $t=1$ to $T$, we get
 \begin{align}\label{conver-core}
\frac{1}{T}\sum_{t=1}^T\EE\|\nabla f(\bm{w}^{t})\|_1&\leq \frac{2(f(\bm{w}^{1})-\min f)}{\gamma T}+  4\sqrt{d}\sum_{t=1}^T\EE\|\bm{\epsilon}^t\|/T
 +L_1\gamma d,
\end{align}
yielding the following result
 \begin{align*}
\frac{1}{T}\sum_{t=1}^T\EE\|\nabla f(\bm{w}^{t})\|_1&\leq \frac{2(f(\bm{w}^{1})-\min f)}{\gamma T}+
 L_1\gamma d+\frac{4\sqrt{d}L_1\gamma}{1-\theta} \\
 &\qquad+\frac{4\sqrt{d}\|\nabla f(\bm{w}^0)\|}{(1-\theta)T}+\frac{4L_2\sqrt{d}\gamma}{(1-\theta)}\frac{\sum_{t=1}^{T}\EE\|\nabla f(\bm{w}^{t})\|}{T}+8\sqrt{d}\sqrt{1-\theta}\sigma\\
 &\leq \frac{2(f(\bm{w}^{1})-\min f)}{\gamma T}+
 L_1\gamma d+\frac{4\sqrt{d}L_1\gamma}{1-\theta} \\
 &\qquad+\frac{4\sqrt{d}\|\nabla f(\bm{w}^0)\|}{(1-\theta)T}+\frac{4L_2\sqrt{d}\gamma}{(1-\theta)}\frac{\sum_{t=1}^{T}\EE\|\nabla f(\bm{w}^{t})\|_1}{T}+8\sqrt{d}\sqrt{1-\theta}\sigma.
\end{align*}
Setting   $1-\theta=\frac{1}{\sqrt{T}}, \gamma=\frac{1}{L_1T^{3/4}}$,
 \begin{align*}
(1-\frac{4L_2\sqrt{d}}{L_1T^{1/4}})\frac{1}{T}\sum_{t=1}^T\EE\|\nabla f(\bm{w}^{t})\|_1&\leq \frac{2L_1(f(\bm{w}^{1})-\min f)}{T^{1/4}}+
\frac{d}{T^{3/4}}+\frac{4\sqrt{d}}{T^{1/4}} \\
 &\quad+\frac{4\sqrt{d}\|\nabla f(\bm{w}^0)\|}{\sqrt{T}}+\frac{8\sqrt{d}\sigma}{T^{1/4}}.
\end{align*}
As $T\geq \frac{4096d^2L_2^4}{L_1^4}$, it holds
 \begin{align*}
\frac{1}{T}\sum_{t=1}^T\EE\|\nabla f(\bm{w}^{t})\|_1&\leq \frac{4L_1(f(\bm{w}^{1})-\min f)}{T^{1/4}}+
\frac{2d}{T^{3/4}}+\frac{8\sqrt{d}}{T^{1/4}} \\
 &\quad+\frac{8\sqrt{d}\|\nabla f(\bm{w}^0)\|}{\sqrt{T}}+\frac{16\sqrt{d}\sigma}{T^{1/4}}.
\end{align*}

\noindent\textbf{Case 2: $\zeta=\frac{\theta}{1-\theta}$}.  We adopt the following notation:
\begin{equation}\label{th2-nota}
\begin{aligned}
\hat{\bm{g}}^t:=\nabla f(\bm{v}^t;\xi^t), \bm{\epsilon}^t:=\bm{m}^t-\nabla f(\bm{w}^t),\hat{\bm{\delta}}^t:=\hat{\bm{g}}^t-\nabla f(\bm{v}^t),\\
{\bf H}({\bf x},\bm{y}):=\nabla f(\bm{y})-\nabla f(\bm{x})-[\nabla^2 f(\bm{x})](\bm{y}-\bm{x}).
\end{aligned}
\end{equation}
From the  notation \eqref{th2-nota}, we derive
\begin{align*}
\bm{m}^{t}&=\theta\bm{m}^{t-1}+(1-\theta)\hat{\bm{g}}^t\\
&=\theta(\bm{\epsilon}^{t-1}+\nabla f(\bm{w}^{t-1}))+(1-\theta)(\hat{\bm{\delta}}^t+\nabla f(\bm{v}^t))\\
&=\theta\Big[\bm{\epsilon}^{t-1}+\nabla f(\bm{w}^{t})+[\nabla^2 f(\bm{w}^{t})](\bm{w}^{t-1}-\bm{w}^{t})+{ \bf H}(\bm{w}^{t-1},\bm{w}^{t})\Big]\\
&\quad\qquad+(1-\theta)\Big[\hat{\bm{\delta}}^t+\nabla f(\bm{w}^{t})+[\nabla^2 f(\bm{w}^{t})](\bm{v}^{t}-\bm{w}^{t})+{\bf H}(\bm{v}^{t},\bm{w}^{t})\Big]\\
&=\theta\Big[\bm{\epsilon}^{t-1}+\nabla f(\bm{w}^{t})+{\bf H}(\bm{w}^{t-1},\bm{w}^{t})\Big]+(1-\theta)\Big[\hat{\bm{\delta}}^t+\nabla f(\bm{w}^{t})+{\bf H}(\bm{v}^{t},\bm{w}^{t})\Big],
\end{align*}
where the last equality used $\bm{v}^{t}-\bm{w}^{t}=\frac{\theta}{1-\theta}(\bm{w}^{t}-\bm{w}^{t-1})$.
Subtracting both sides with $\nabla f(\bm{w}^t)$,
\begin{align*}
\bm{\epsilon}^t=\bm{m}^{t}-\nabla f(\bm{w}^t)=\theta\bm{\epsilon}^{t-1}+\theta{\bf H}(\bm{w}^{t-1},\bm{w}^{t})+(1-\theta){\bf H}(\bm{v}^{t},\bm{w}^{t})+(1-\theta)\hat{\bm{\delta}}^t.
\end{align*}
The mathematical induction method gives us
\begin{align*}
  \bm{\epsilon}^t= \theta^t\bm{\epsilon}^{0}+ \theta\sum_{i=1}^{t}\theta^{t-i}{\bf H}(\bm{w}^{i-1},\bm{w}^{i})+(1-\theta)\sum_{i=1}^{t}\theta^{t-i}{\bf H}(\bm{v}^{i},\bm{w}^{i})+(1-\theta)\sum_{i=1}^{t}\theta^{t-i}\hat{\bm{\delta}}^i.
\end{align*}
Taking the norms and expectations of both sides of the above equation and noticing that $\gamma\leq R$, we can derive
\begin{align}\label{important}
\begin{aligned}
 \EE\| \bm{\epsilon}^t\|&\leq   \theta\sum_{i=1}^{t}\theta^{t-i}\EE\|{\bf H}(\bm{w}^{i-1},\bm{w}^{i})\|+(1-\theta)\sum_{i=1}^{t}\theta^{t-i}\EE\|{\bf H}(\bm{v}^{i},\bm{w}^{i})\|\\
 &\qquad+\EE\left\|(1-\theta)\sum_{i=1}^{t}\theta^{t-i}\hat{\bm{\delta}}^i\right\|+\theta^t\|\bm{\epsilon}^{0}\|\\
 &\leq\theta\sum_{i=1}^{t}\theta^{t-i}\EE\frac{(H_1+H_2\|\nabla  f(\bm{w}^{i})\|)}{2}\|\bm{w}^{i-1}-\bm{w}^{i}\|^2\\
 &\qquad+(1-\theta)\sum_{i=1}^{t}\theta^{t-i}\EE\frac{(H_1+H_2\|\nabla  f(\bm{w}^{i})\|)}{2}\|\bm{v}^{i}-\bm{w}^{i}\|^2\\
 &\qquad+\EE\left\|(1-\theta)\sum_{i=1}^{t}\theta^{t-i}\hat{\bm{\delta}}^i\right\|+\theta^t\|\bm{\epsilon}^{0}\|\\
 &\overset{\bm{v}^{i}-\bm{w}^{i}=\frac{\theta}{1-\theta}(\bm{w}^{i}-\bm{w}^{i-1})}{\leq}\frac{\theta}{1-\theta}\sum_{i=1}^{t}\theta^{t-i}\EE\frac{(H_1+H_2\|\nabla  f(\bm{w}^{i})\|)}{2}\|\bm{w}^{i-1}-\bm{w}^{i}\|^2\\
 &\qquad+\EE \left\|(1-\theta)\sum_{i=1}^{t}\theta^{t-i}\hat{\bm{\delta}}^i \right\|+\theta^t\|\bm{\epsilon}^{0}\|,
\end{aligned}
\end{align}
where we used $
\|{\bf H}(\bm{w}^{i-1},\bm{w}^{i})\|\leq\frac{(H_1+H_2\|\nabla  f(\bm{w}^{i})\|)}{2}\|\bm{w}^{i-1}-\bm{w}^{i}\|^2
$ and $
\|{\bf H}(\bm{v}^{i},\bm{w}^{i})\|\leq\frac{(H_1+H_2\|\nabla  f(\bm{w}^{i})\|)}{2}\|\bm{v}^{i}-\bm{w}^{i}\|^2
$ from Lemma \ref{lemma0}.
Recalling \eqref{important} with
$$\|\bm{w}^{i-1}-\bm{w}^{i}\|^2=\|\gamma \SI(\bm{m}^{i-1})\|^2=\gamma^2d,$$
we obtain the following bound for $\EE\| \bm{\epsilon}^t\|$  as
\begin{align}\label{sign-sec}
 \EE\| \bm{\epsilon}^t\|\leq \frac{H_1}{2}\frac{\theta}{(1-\theta)^2}\gamma^2d+\frac{H_2}{2}\frac{\theta}{1-\theta}\gamma^2d\sum_{i=1}^t\theta^{t-i}\EE\|\nabla  f(\bm{w}^{i})\|+\sqrt{1-\theta}\sigma+\theta^t\|\bm{\epsilon}^{0}\|.
\end{align}
Using Lemma \ref{lemma1-sign} by setting  $\bm{x}^{\dag}\rightarrow \bm{x}^{t},\bm{m}\rightarrow \bm{m}^{t}$ and taking expectations, as $\gamma\leq  r$
\begin{align}\label{sign-ineq}
\EE f(\bm{w}^{t+1})-\EE f(\bm{w}^{t})\leq -\gamma/2 \EE\|\nabla f(\bm{w}^{t})\|_1+2\sqrt{d}\gamma\EE\|\bm{\epsilon}^t\|+\frac{L_1\gamma^2}{2}d.
\end{align}
By summing   inequality \eqref{sign-ineq} with $t$ ranging from  $1$ to $T$ and  noticing $0\leq\theta\leq 1$, we have
\begin{align*}
\frac{1}{T}\sum_{t=1}^T\EE\|\nabla f(\bm{w}^{t})\|_1&\leq \frac{2(f(\bm{w}^{1})-\min f)}{\gamma T} +\frac{2H_1}{(1-\theta)^2}\gamma^2d^{3/2}+\frac{2H_2}{(1-\theta)^2}\gamma^2d^{3/2}\frac{1}{T}\sum_{t=1}^T\EE\|\nabla  f(\bm{w}^{t})\|\\
&+4\sqrt{1-\theta}\sigma+2L_1\gamma d+4\sqrt{d}\sum_{t=1}^T\theta^t\|\bm{\epsilon}^{0}\|/T\\
\Longrightarrow &\Big[1-\frac{2H_2}{(1-\theta)^2}\gamma^2d^{3/2}\Big]\frac{1}{T}\sum_{t=1}^T\EE\|\nabla f(\bm{w}^{t})\|_1\leq \frac{2(f(\bm{w}^{1})-\min f)}{\gamma T} +\frac{2H_1}{(1-\theta)^2}\gamma^2d^{3/2}\\
&+4\sqrt{1-\theta}\sigma+2L_1\gamma d+4\sqrt{d}\sum_{t=1}^T\theta^t\|\bm{\epsilon}^{0}\|/T,
\end{align*}
where we used the fact $\|\cdot\|\leq\|\cdot\|_1$.
By setting  $1-\theta=\frac{1}{T^{4/7}}$, $\gamma=\frac{1}{\max\{\sqrt{H_1},\sqrt{H_2},L_1\}T^{5/7}}$, and the fact $\bm{w}^0=\bm{w}^1$, we get
\begin{align*}
 \Big[1-\frac{2d^{3/2}}{T^{2/7}}\Big]\frac{1}{T}\sum_{t=1}^T\EE\|\nabla f(\bm{w}^{t})\|_1&\leq \frac{2\max\{\sqrt{H_1},\sqrt{H_2},L_1\}(f(\bm{w}^{0})-\min f)}{ T^{2/7}} +\frac{2d^{3/2}}{ T^{2/7}}\\
&+\frac{4\sigma}{T^{2/7}}+\frac{2d}{ T^{5/7}}+\frac{4\sqrt{d}\|\nabla f(\bm{w}^0)\|}{T^{3/7}}.
\end{align*}
As $T\geq(4d^{3/2})^{7/2}=2^7d^{21/4}$, $1-\frac{2}{T^{2/7}}\geq\frac{1}{2}$. Then, we get
\begin{align*}
\frac{1}{T}\sum_{t=1}^T\EE\|\nabla f(\bm{w}^{t})\|_1&\leq \frac{4\max\{\sqrt{H_1},\sqrt{H_2},L_1\}(f(\bm{w}^{0})-\min f)}{T^{2/7}} +\frac{4d^{3/2}}{T^{2/7}}\\
&+\frac{8\sigma}{T^{2/7}}+\frac{4d}{T^{5/7}}+\frac{8\sqrt{d}\|\nabla f(\bm{w}^0)\|}{T^{3/7}}.
\end{align*}
Due to that Lemma \ref{lemma1-sign} requires $\gamma=\frac{1}{\max\{\sqrt{H_1},\sqrt{H_2},L_1\}T^{5/7}}\leq\max\{r,R,\frac{1}{dL_2}\}$, giving that
$$T\geq \max\{ (\frac{L_2}{\max\{\sqrt{H_1},\sqrt{H_2},L_1\}})^{\frac{7}{5}},
\frac{1}{(r\max\{\sqrt{H_1},\sqrt{H_2},L_1\})^{\frac{7}{5}}},
\frac{1}{(R\max\{\sqrt{H_1},\sqrt{H_2},L_1\})^{\frac{7}{5}}}\}.$$
In summary, we have
$$T\geq\max\{ (\frac{dL_2}{\max\{\sqrt{H_1},\sqrt{H_2},L_1\}})^{\frac{7}{5}},
\frac{1}{(r\max\{\sqrt{H_1},\sqrt{H_2},L_1\})^{\frac{7}{5}}},
\frac{1}{(R\max\{\sqrt{H_1},\sqrt{H_2},L_1\})^{\frac{7}{5}}},2^7d^{21/4}\}.$$
\section{Proof of Theorem \ref{th3}}
\textbf{Case 1: $\zeta=0$}.
We consider the short hand notation as follows
\begin{equation}\label{th1-reuse-n}
\bm{g}^t:=\sum_{i=1}^n\hat{\bm{g}}^t(i)/n, \bm{\epsilon}^t:=\bm{m}^t-\nabla f(\bm{w}^t), \bm{\delta}^t:=\bm{g}^t-\nabla f(\bm{w}^t),\bm{s}^t:=\nabla f(\bm{w}^{t-1})-\nabla f(\bm{w}^{t}).
\end{equation}
The scheme of algorithm gives us
\begin{align*}
\bm{m}^{t}=\theta\bm{m}^{t-1}+(1-\theta)\bm{g}^t=\theta\Big[\bm{\epsilon}^{t-1}+\nabla f(\bm{w}^{t-1})\Big]+(1-\theta)\Big[\bm{\delta}^t+\nabla f(\bm{w}^t)\Big],
\end{align*}
which also yields
\begin{align*}
\bm{\epsilon}^t=\bm{m}^{t}-\nabla f(\bm{w}^t)=\theta\bm{\epsilon}^{t-1}+\theta\bm{s}^t+(1-\theta)\bm{\delta}^t.
\end{align*}
With Mathematical Induction, we can further get
\begin{align}\label{ada-reuse}
  \bm{\epsilon}^t= \theta^t\bm{\epsilon}^{0}+\sum_{k=1}^{t}\theta^{t-k}\bm{s}^k+(1-\theta)\sum_{k=1}^{t}\theta^{t-k}\bm{\delta}^k.
\end{align}
Taking the norms of both sides of inequality~\eqref{ada-reuse},
\begin{align*}
   \EE\|\bm{\epsilon}^t\|\leq \sum_{k=1}^{t}\theta^{t-k}\EE\|\bm{s}^k\|+(1-\theta)\EE\Big\|\sum_{k=1}^{t}\theta^{t-k}\bm{\delta}^k\Big\|+\theta^t\|\bm{\epsilon}^{0}\|.
\end{align*}
As $\gamma\leq r$, the weak smooth property of function $f$ tells us
\begin{align*}
\EE\|\bm{s}^k\|\leq \EE\Big[(L_1+L_2\|\nabla f(\bm{w}^{k-1})\|)\|\bm{w}^{k}-\bm{w}^{k-1}\|\Big]=L_1\sqrt{d}\gamma+L_2\sqrt{d}\gamma\EE\|\nabla f(\bm{w}^{k-1})\|.
\end{align*}
Turning back to \eqref{ada-reuse}, we derive
\begin{align}\label{ada-reuse2}
  \EE\|\bm{\epsilon}^t\|\leq L_1\sqrt{d}\gamma\sum_{k=1}^{t}\theta^{t-k}+L_2\sqrt{d}\gamma\sum_{k=1}^{t}\theta^{t-k}\EE\|\nabla f(\bm{w}^{k-1})\|+(1-\theta)\EE\Big\|\sum_{k=1}^{t}\theta^{t-k}\bm{\delta}^k\Big\|+\theta^t\|\bm{\epsilon}^{0}\|.
\end{align}
 Thus, we just need to bound $\EE\Big\|\sum_{k=1}^{t}\theta^{t-k}\bm{\delta}^k\Big\|^2$.
The Cauchy's inequality yields
\begin{align}\label{th3-t1}
\begin{aligned}
(1-\theta)\EE_{\CC,\chi^t}\Big\|\sum_{k=1}^{t}\theta^{t-k}\bm{\delta}^k\Big\|&\leq (1-\theta)\sqrt{\EE_{\CC,\chi^t}\Big\|\sum_{k=1}^{t}\theta^{t-k}\bm{\delta}^k\Big\|^2}\\
&=(1-\theta)\sqrt{\sum_{k=1}^{t}\theta^{2t-2k}\EE_{\CC,\chi^t}\|\bm{\delta}^k\|^2+2\sum_{k< j\leq t}\theta^{2t-(k+j)}\EE_{\CC,\chi^t}\langle\bm{\delta}^k,\bm{\delta}^j\rangle}.
\end{aligned}
\end{align}
When $\zeta=0$, for any $t\in \mathbb{Z}^+$, it holds
$$\bm{w}^t=\bm{v}^t.$$
With Lemma \ref{lemma2}, we have
\begin{align*}
\sum_{k=1}^{t}\theta^{2t-2k}\EE_{\CC,\chi^t}\|\bm{\delta}^k\|^2&\leq\sum_{k=1}^{t}\theta^{2t-2k}\Big[(2\sigma^2+4\bar{\sigma}^2)(1-\delta)^u+\frac{2\sigma^2}{n}+4(1-\delta)^u\|\nabla f(\bm{w}^k)\|^2\Big]\\
&\leq\frac{1}{1-\theta^2}\Big[(2\sigma^2+4\bar{\sigma}^2)(1-\delta)^u+\frac{2\sigma^2}{n}\Big]+4(1-\delta)^u\sum_{k=1}^{t}\theta^{2t-2k}\|\nabla f(\bm{w}^k)\|^2\\
&\leq\frac{1}{1-\theta^2}\Big[(2\sigma^2+4\bar{\sigma}^2)(1-\delta)^u+\frac{2\sigma^2}{n}\Big]+4(1-\delta)^u\sum_{k=1}^{t}\theta^{t-k}\|\nabla f(\bm{w}^k)\|^2
\end{align*}
On the other hand, with Lemma \ref{lemma3},
\begin{align*}
&2\sum_{i< k\leq t}\theta^{2t-(k+j)}\EE_{\CC,\chi^t}\langle\bm{\delta}^k,\bm{\delta}^j\rangle\\
&\leq\sum_{k< j\leq t}\theta^{2t-(k+j)}\Big[(2\sigma^2+4\bar{\sigma}^2)(1-\delta)^u+2(1-\delta)^u\|\nabla f(\bm{w}^k)\|^2+2(1-\delta)^u\|\nabla f(\bm{w}^j)\|^2\Big]\\
&\leq \frac{1}{1-\theta}\sum_{k=1}^{t}\theta^{t-k}\Big[(2\sigma^2+4\bar{\sigma}^2)(1-\delta)^u+4(1-\delta)^u\|\nabla f(\bm{w}^k)\|^2\Big]\\
&\leq \frac{2\sigma^2+4\bar{\sigma}^2}{(1-\theta)^2}(1-\delta)^u+\frac{1}{1-\theta}\sum_{k=1}^{t}\theta^{t-k}\Big[4(1-\delta)^u\|\nabla f(\bm{w}^k)\|^2\Big]
\end{align*}
Turning back to \eqref{th3-t1},
\begin{align}\label{th3-t2}
\begin{aligned}
&(1-\theta)\EE_{\CC,\chi^t}\Big\|\sum_{k=1}^{t}\theta^{t-k}\bm{\delta}^k\Big\|\\
&\leq(1-\theta)\sqrt{\frac{2\sigma^2}{(1-\theta^2)n}+\frac{(2\sigma^2+4\bar{\sigma}^2)(1-\delta)^u}{(1-\theta^2)}+\frac{(2\sigma^2+4\bar{\sigma}^2)(1-\delta)^u}{(1-\theta)^2}+\frac{8(1-\delta)^u}{(1-\theta)}\sum_{k=1}^{t}\theta^{t-k}\|\nabla f(\bm{w}^k)\|^2}\\
&\leq\sqrt{1-\theta}\frac{2\sigma}{\sqrt{n}}+\sqrt{1-\theta}\sqrt{2\sigma^2+4\bar{\sigma}^2}(1-\delta)^{u/2}+\sqrt{2\sigma^2+4\bar{\sigma}^2}(1-\delta)^{u/2}\\
&\qquad+2\sqrt{2}\sqrt{1-\theta}(1-\delta)^{u/2}\sum_{k=1}^{t}\theta^{\frac{t-k}{2}}\|\nabla f(\bm{w}^k)\|
\end{aligned}
\end{align}
where we used $\sqrt{a+b}\leq\sqrt{a}+\sqrt{b}$. Noticing that $\EE\Big(\EE_{\CC,\chi^t}\Big\|\sum_{k=1}^{t}\theta^{t-k}\bm{\delta}^k\Big\|\Big)=\EE\Big\|\sum_{k=1}^{t}\theta^{t-k}\bm{\delta}^k\Big\|$, we have
\begin{equation}\label{wucha-dis-pre}
\begin{aligned}
&(1-\theta)\EE\Big\|\sum_{k=1}^{t}\theta^{t-k}\bm{\delta}^k\Big\|\\
&\leq\sqrt{1-\theta}\frac{2\sigma}{\sqrt{n}}+\sqrt{1-\theta}\sqrt{2\sigma^2+4\bar{\sigma}^2}(1-\delta)^{u/2}+\sqrt{2\sigma^2+4\bar{\sigma}^2}(1-\delta)^{u/2}\\
&\qquad+2\sqrt{2}\sqrt{1-\theta}(1-\delta)^{u/2}\sum_{k=1}^{t}\theta^{\frac{t-k}{2}}\EE\|\nabla f(\bm{w}^k)\|.
\end{aligned}
\end{equation}
Therefore, we are led to
\begin{align}\label{wucha-dis}
\begin{aligned}
  \sum_{t=1}^{T}\EE\|\bm{\epsilon}^t\|/T
 &\leq\frac{ L_1\sqrt{d}\gamma}{1-\theta} +\frac{\|\nabla f(\bm{w}^0)\|}{(1-\theta)T}+\frac{L_2\sqrt{d}\gamma}{(1-\theta)}\frac{\sum_{t=1}^{T}\EE\|\nabla f(\bm{w}^{t})\|}{T}\\
 &\quad+\sqrt{1-\theta}\frac{2\sigma}{\sqrt{n}}+\sqrt{1-\theta}\sqrt{2\sigma^2+4\bar{\sigma}^2}(1-\delta)^{u/2}+\sqrt{2\sigma^2+4\bar{\sigma}^2}(1-\delta)^{u/2}\\
&\quad+\frac{4\sqrt{2}}{\sqrt{1-\theta}}(1-\delta)^{u/2}\frac{\sum_{t=1}^{T}\EE\|\nabla f(\bm{w}^{t})\|}{T},
 \end{aligned}
\end{align}
where we used that $\sum_{t=1}^T\sum_{k=1}^{t}\theta^{\frac{t-k}{2}}\EE\|\nabla f(\bm{w}^k)\|/T\leq\frac{1}{1-\sqrt{\theta}}\frac{\sum_{t=1}^{T}\EE\|\nabla f(\bm{w}^{t})\|}{T}$ and $\frac{\sqrt{1-\theta}}{1-\sqrt{\theta}}=\frac{1+\sqrt{\theta}}{\sqrt{1-\theta}}\leq\frac{2}{\sqrt{1-\theta}}$.
Using Lemma \ref{lemma1-sign} with  $\bm{w}^{\dag}\rightarrow \bm{w}^{t}$ and $\bm{m}\rightarrow \bm{m}^{t}$ and taking taking expectations,
\begin{align*}
\EE (f(\bm{w}^{t+1}))-\EE f(\bm{w}^{t})\leq -\gamma/2 \EE\|\nabla f(\bm{w}^{t})\|+2\sqrt{d}\gamma\EE\|\bm{\epsilon}^t\|+\frac{L_1}{2}\gamma^2d.
\end{align*}
Summing the recursion from $t=1$ to $T$, we get
 \begin{align}\label{conver-core}
\frac{1}{T}\sum_{t=1}^T\EE\|\nabla f(\bm{w}^{t})\|&\leq \frac{2(f(\bm{w}^{1})-\min f)}{\gamma T}+  4\sqrt{d}\sum_{t=1}^T\EE\|\bm{\epsilon}^t\|/T
 +L_1\gamma d.
\end{align}
Based on the inequality, we get
 \begin{align*}
\frac{1}{T}\sum_{t=1}^T\EE\|\nabla f(\bm{w}^{t})\|&\leq \frac{2(f(\bm{w}^{1})-\min f)}{\gamma T}+
 L_1\gamma d\\
 &\quad+\frac{4L_1d\gamma}{1-\theta} +\frac{4\sqrt{d}\|\nabla f(\bm{w}^0)\|}{(1-\theta)T}+\frac{4L_2d\gamma}{(1-\theta)}\frac{\sum_{t=1}^{T}\EE\|\nabla f(\bm{w}^{t})\|}{T}\\
 &\quad+8\sqrt{1-\theta}\frac{\sqrt{d}\sigma}{\sqrt{n}}+4\sqrt{d}\sqrt{1-\theta}\sqrt{2\sigma^2+4\bar{\sigma}^2}(1-\delta)^{u/2}+4\sqrt{d}\sqrt{2\sigma^2+4\bar{\sigma}^2}(1-\delta)^{u/2}\\
&\quad+\frac{16\sqrt{2d}}{\sqrt{1-\theta}}(1-\delta)^{u/2}\frac{\sum_{t=1}^{T}\EE\|\nabla f(\bm{w}^{t})\|}{T}.
\end{align*}
By setting $1-\theta=\frac{\sqrt{n}}{\sqrt{T}}$ and $\gamma=\frac{n^{1/4}}{L_1T^{3/4}}$,
 \begin{align*}
&\Big(1-\frac{4dL_2}{L_1T^{1/4}n^{1/4}}-\frac{16\sqrt{2d}T^{1/4}}{n^{1/4}}(1-\delta)^{u/2}\Big)\frac{1}{T}\sum_{t=1}^T\EE\|\nabla f(\bm{w}^{t})\|\\
&\leq \frac{2L_1(f(\bm{w}^{1})-\min f)}{n^{1/4}T^{1/4}}+\frac{8d}{T^{1/4}n^{1/4}} +\frac{4\sqrt{d}\|\nabla f(\bm{w}^0)\|}{\sqrt{nT}}+\frac{8\sqrt{d}\sigma}{T^{1/4}n^{1/4}}\\
 &\qquad+\frac{4\sqrt{d}\sqrt{2\sigma^2+4\bar{\sigma}^2}}{T^{1/4}}(1-\delta)^{u/2}+4\sqrt{d}\sqrt{2\sigma^2+4\bar{\sigma}^2}(1-\delta)^{u/2}\\
 &\leq \frac{2L_1(f(\bm{w}^{1})-\min f)}{n^{1/4}T^{1/4}}+\frac{8d}{n^{1/4}T^{1/4}} +\frac{4\sqrt{d}\|\nabla f(\bm{w}^0)\|}{\sqrt{nT}}+\frac{8\sqrt{d}\sigma}{n^{1/4}T^{1/4}}+8\sqrt{d}\sqrt{2\sigma^2+4\bar{\sigma}^2}(1-\delta)^{u/2}
\end{align*}
where we used $1\leq\frac{1}{1-\theta}$. As $\frac{L_2}{L_1T^{1/4}}\leq\frac{1}{16d}$ and $16\sqrt{2d}T^{1/4}(1-\delta)^{u/2}=\frac{1}{4n}$, i.e.,
$$T\geq(\frac{16dL_2}{L_1})^4,~~u=\frac{2\ln(64\sqrt{2d}nT^{1/4})}{\ln(\frac{1}{1-\delta})},$$
we have
$$1-\frac{4dL_2}{L_1T^{1/4}n^{1/4}}-\frac{16\sqrt{2d}T^{1/4}}{n^{1/4}}(1-\delta)^{u/2}\geq \frac{3}{4}-\frac{1}{4n^{5/4}}\geq\frac{1}{2},$$
also yielding
 \begin{align*}
\frac{1}{T}\sum_{t=1}^T\EE\|\nabla f(\bm{w}^{t})\|\leq \frac{4L_1(f(\bm{w}^{1})-\min f)}{n^{1/4}T^{1/4}}+\frac{16d}{n^{1/4}T^{1/4}} +\frac{8\sqrt{d}\|\nabla f(\bm{w}^0)\|}{\sqrt{nT}}+\frac{16\sqrt{d}\sigma}{n^{1/4}T^{1/4}}+\frac{\sqrt{\sigma^2+2\bar{\sigma}^2}}{8T^{1/4}n}.
\end{align*}
Because Lemma \ref{lemma1-sign} requires $\gamma=\frac{1}{L_1T^{3/4}}\leq\max\{\frac{1}{L_2d},r\}$, indicating $T\geq \max\{(\frac{L_2d}{L_1})^{\frac{4}{3}},\frac{1}{(L_1r)^{\frac{4}{3}}}\}$.
With the fact $\bm{w}^{1}=\bm{w}^{0}$, we then proved the result.
\medskip\\

\noindent\textbf{Case 2: $\zeta=\frac{\theta}{1-\theta}$}.  Following the same notation in the proof of Theorem \ref{th4}, i.e., \eqref{th2-nota}. Noticing that \eqref{important} still holds, we just need to bound $\EE\Big\|\sum_{k=1}^{t}\theta^{t-k}\hat{\bm{\delta}}^k\Big\|^2$.
Similar to the proof of Case 1, we can just need to replace $\EE\|\nabla f(\bm{w}^k)\|$ with $\EE\|\nabla f(\bm{v}^k)\|$ in bound of \eqref{wucha-dis-pre}, i.e.,
\begin{align*}
\begin{aligned}
&(1-\theta)\EE\Big\|\sum_{k=1}^{t}\theta^{t-k}\hat{\bm{\delta}}^k\Big\|\\
&\leq\sqrt{1-\theta}\frac{\sigma}{\sqrt{n}}+\sqrt{1-\theta}\sqrt{2\sigma^2+4\bar{\sigma}^2}(1-\delta)^{u/2}+\sqrt{2\sigma^2+4\bar{\sigma}^2}(1-\delta)^{u/2}\\
&\qquad+2\sqrt{2}\sqrt{1-\theta}(1-\delta)^{u/2}\sum_{k=1}^{t}\theta^{\frac{t-k}{2}}\EE\|\nabla f(\bm{v}^k)\|.
\end{aligned}
\end{align*}
With the weak first-order Lipschitz property,
\begin{align*}
\EE\|\nabla f(\bm{v}^k)\|&=\EE\|\nabla f(\bm{v}^k)-\nabla f(\bm{w}^k)+\nabla f(\bm{w}^k)\|\\
&\leq\EE\|\nabla f(\bm{v}^k)-\nabla f(\bm{w}^k)\|+\EE\|\nabla f(\bm{w}^k)\|\\
&\leq\EE\Big[(L_1+L_2\|\nabla f(\bm{w}^k)\|)\|\bm{v}^k-\bm{v}^k\|\Big]+\EE\|\nabla f(\bm{w}^k)\|\\
&\leq \EE\Big[(L_1+L_2\|\nabla f(\bm{w}^k)\|)\|\frac{\theta}{1-\theta}(\bm{w}^{k}-\bm{w}^{k-1})\|\Big]+\EE\|\nabla f(\bm{w}^k)\|\\
&\leq \frac{\gamma\sqrt{d}}{1-\theta}\EE\Big[(L_1+L_2\|\nabla f(\bm{w}^k)\|)\Big]+\EE\|\nabla f(\bm{w}^k)\|\\
&\leq \frac{\gamma\sqrt{d} L_1}{1-\theta}+(\frac{\gamma\sqrt{d} L_2}{1-\theta}+1)\EE\|\nabla f(\bm{w}^k)\|
\end{align*}
With this bound, we get
\begin{align}\label{wucha-dis-pre2}
\begin{aligned}
&(1-\theta)\EE\Big\|\sum_{k=1}^{t}\theta^{t-k}\hat{\bm{\delta}}^k\Big\|\\
&\leq\sqrt{1-\theta}\frac{2\sigma}{\sqrt{n}}+\sqrt{1-\theta}\sqrt{2\sigma^2+4\bar{\sigma}^2}(1-\delta)^{u/2}+\sqrt{2\sigma^2+4\bar{\sigma}^2}(1-\delta)^{u/2}\\
&\qquad+2\sqrt{2}\sqrt{1-\theta}(1-\delta)^{u/2}\sum_{k=1}^{t}\theta^{\frac{t-k}{2}}(\frac{\gamma \sqrt{d}L_1}{1-\theta}+(\frac{\gamma\sqrt{d} L_2}{1-\theta}+1)\EE\|\nabla f(\bm{w}^k)\|).
\end{aligned}
\end{align}
Replacing the term  $\sqrt{1-\theta}\sigma$ with the bound \eqref{wucha-dis-pre2} in \eqref{sign-sec},
\begin{align*}
 \EE\| \bm{\epsilon}^t\|&\leq \frac{H_1}{2}\frac{\theta}{(1-\theta)^2}\gamma^2d+\frac{H_2}{2}\frac{\theta}{1-\theta}\gamma^2d\sum_{k=1}^t\theta^{t-k}\EE\|\nabla  f(\bm{w}^{k})\|+\theta^t\|\bm{\epsilon}^{0}\|\\
 &\qquad+\sqrt{1-\theta}\frac{2\sigma}{\sqrt{n}}+\sqrt{1-\theta}\sqrt{2\sigma^2+4\bar{\sigma}^2}(1-\delta)^{u/2}+\sqrt{2\sigma^2+4\bar{\sigma}^2}(1-\delta)^{u/2}\\
&\qquad+2\sqrt{2}\sqrt{1-\theta}(1-\delta)^{u/2}\sum_{k=1}^{t}\theta^{\frac{t-k}{2}}(\frac{\gamma\sqrt{d} L_1}{1-\theta}+(\frac{\gamma\sqrt{d} L_2}{1-\theta}+1)\EE\|\nabla f(\bm{w}^k)\|).
\end{align*}
Summing the inequality from $t=1$ to $T$,
\begin{align*}
 \sum_{t=1}^T\EE\| \bm{\epsilon}^t\|/T&\leq \frac{H_1}{2}\frac{\theta}{(1-\theta)^2}\gamma^2d+\frac{H_2}{2}\frac{\theta}{(1-\theta)^2}\gamma^2d\frac{\sum_{t=1}^T\EE\|\nabla  f(\bm{w}^{t})\|}{T}+ \frac{\|\bm{\epsilon}^{0}\|}{(1-\theta)T}\\
 &\qquad+\sqrt{1-\theta}\frac{2\sigma}{\sqrt{n}}+\sqrt{1-\theta}\sqrt{2\sigma^2+4\bar{\sigma}^2}(1-\delta)^{u/2}+\sqrt{2\sigma^2+4\bar{\sigma}^2}(1-\delta)^{u/2}\\
&\qquad+\frac{4\sqrt{2}}{\sqrt{1-\theta}}(1-\delta)^{u/2}(\frac{\gamma\sqrt{d} L_2}{1-\theta}+1)\frac{\sum_{t=1}^{T}\EE\|\nabla f(\bm{w}^{t})\|}{T}+\frac{4\sqrt{2}}{\sqrt{1-\theta}}(1-\delta)^{u/2}(\frac{\gamma\sqrt{d} L_1}{1-\theta}).
\end{align*}
In this case, \eqref{conver-core} still holds, and we have
 \begin{align*}
\frac{1}{T}\sum_{t=1}^T\EE\|\nabla f(\bm{w}^{t})\|&\leq \frac{2(f(\bm{w}^{1})-\min f)}{\gamma T}+
 L_1\gamma d\\
 &\quad+2\sqrt{d}H_1\frac{\theta}{(1-\theta)^2}\gamma^2+2H_2\frac{\theta}{(1-\theta)^2}\gamma^2d^{3/2}\frac{\sum_{t=1}^T\EE\|\nabla  f(\bm{w}^{t})\|}{T}+ \frac{4\sqrt{d}\|\bm{\epsilon}^{0}\|}{(1-\theta)T}\\
 &+\sqrt{1-\theta}\frac{8\sqrt{d}\sigma}{\sqrt{n}}+4\sqrt{d}\sqrt{1-\theta}\sqrt{2\sigma^2+4\bar{\sigma}^2}(1-\delta)^{u/2}+4\sqrt{d}\sqrt{2\sigma^2+4\bar{\sigma}^2}(1-\delta)^{u/2}\\
&\qquad+\frac{16\sqrt{2}}{\sqrt{1-\theta}}(1-\delta)^{u/2}(\frac{\gamma dL_2}{1-\theta}+\sqrt{d})\frac{\sum_{t=1}^{T}\EE\|\nabla f(\bm{w}^{t})\|}{T}+\frac{16\sqrt{2}}{\sqrt{1-\theta}}(1-\delta)^{u/2}(\frac{\gamma dL_1}{1-\theta}).
\end{align*}
By setting $1-\theta=\frac{n^{4/7}}{T^{4/7}}$, $\gamma=\frac{n^{2/7}}{CT^{5/7}}$ with $C:=\max\{\sqrt{H_1},\sqrt{H_2},L_1\}$,
 \begin{align*}
&(1-\frac{2H_2d^{3/2}}{C^2T^{2/7}n^{4/7}}-T^{2/7}\frac{(16\sqrt{2}+16\sqrt{2d})}{n^{2/7}}(1-\delta)^{u/2})\frac{1}{T}\sum_{t=1}^T\EE\|\nabla f(\bm{w}^{t})\|\\
&\leq \frac{2C(f(\bm{w}^{1})-\min f)}{n^{2/7}T^{2/7}}+\frac{n^{2/7}d}{T^{5/7}}+\frac{2\sqrt{d}}{n^{4/7}T^{2/7}}+ \frac{4\sqrt{d}\|\bm{\epsilon}^{0}\|}{n^{4/7}T^{3/7}}\\
 &\quad+\frac{8\sqrt{d}\sigma}{n^{2/7}T^{2/7}}+\frac{4\sqrt{d}\sqrt{2\sigma^2+4\bar{\sigma}^2}n^{2/7}}{T^{2/7}}(1-\delta)^{u/2}+4\sqrt{d}\sqrt{2\sigma^2
 +4\bar{\sigma}^2}(1-\delta)^{u/2}+\frac{16\sqrt{2}dT^{1/7}}{n^{4/7}}(1-\delta)^{u/2},
\end{align*}
where we used $\frac{\gamma dL_2}{1-\theta}=\frac{dL_2}{CT^{1/7}n^{2/7}}\leq\frac{dL_2}{CT^{1/7}} \leq 1$ as $T\geq (\frac{dL_2}{C})^7$.
If $\frac{2H_2d^{3/2}}{C^2T^{2/7}}\leq \frac{1}{4}$ and $T(16\sqrt{2}+16\sqrt{2d})(1-\delta)^{u/2}=\frac{1}{4n}$, i.e.,
$$T\geq (\frac{8H_2d^{3/2}}{C^2})^{7/2},~u=\frac{2\ln[(64\sqrt{2}+64\sqrt{2d})nT]}{\ln\frac{1}{1-\delta}},$$
it holds
$$1-\frac{2H_2d^{3/2}}{C^2T^{2/7}n^{4/7}}-\frac{T^{2/7}(16\sqrt{2}+16\sqrt{2d})}{n^{2/7}}(1-\delta)^{u/2}\geq\frac{3}{4}-\frac{1}{4n^{2/7}T^{5/7}}\geq\frac{1}{2}.$$
Under this setting, we have
 \begin{align*}
&\frac{1}{T}\sum_{t=1}^T\EE\|\nabla f(\bm{w}^{t})\|\\
&\leq \frac{4C(f(\bm{w}^{1})-\min f)}{n^{2/7}T^{2/7}}+\frac{2n^{2/7}d}{T^{5/7}}+\frac{4\sqrt{d}}{n^{4/7}T^{2/7}}+ \frac{8\sqrt{d}\|\bm{\epsilon}^{0}\|}{n^{4/7}T^{3/7}}+\frac{16\sqrt{d}\sigma}{n^{2/7}T^{2/7}}+\frac{\sqrt{d}\sqrt{\sigma^2+2\bar{\sigma}^2}}{8nT^{2/7}} +\frac{1}{8T^{6/7}n^{11/7}}\\
 &\leq \frac{4C(f(\bm{w}^{1})-\min f)}{n^{2/7}T^{2/7}}+\frac{3n^{2/7}d}{T^{5/7}}+\frac{4\sqrt{d}}{n^{4/7}T^{2/7}}+ \frac{8\sqrt{d}\|\bm{\epsilon}^{0}\|}{n^{4/7}T^{3/7}}+\frac{16\sqrt{d}\sigma}{n^{2/7}T^{2/7}}+\frac{\sqrt{d}\sqrt{\sigma^2+2\bar{\sigma}^2}}{8nT^{2/7}}.
\end{align*}
As $T\geq n^{4}$, we can get
$$\frac{3n^{2/7}}{T^{5/7}}\leq \frac{3}{n^{2/7}T^{4/7}}$$
and then prove the result.

\section{Proofs of Propositions}
This section will use the Gronwall's inequality.
\begin{lemma}[Gronwall's inequality, \cite{gronwall1919note}]\label{lemma-gro}
 Let $I = [a, b]$ denote an interval of the real
line. Let $f, g, h$ be continuous real-valued functions defined on $I$. Assume $g$ is nondecreasing, $h$ is non-negative, and the negative part of $g$ is integrable on every closed and bounded
subinterval of $I$. If
$f(t)\leq g(t) +\int_{a}^t h(s)f(s)ds, \forall t\in I$,
then
$$f(t)\leq g(t)\exp(\int_{a}^t h(s)ds).$$
\end{lemma}
\subsection{Proof of Proposition \ref{pro1}}
\textbf{About Assumption \ref{ass1}:}
With direct calculations, it holds
$$\nabla D(\bm{x})=(\bm{x}\bm{x}^{\top}-\bm{Y})\bm{x}\in\RR^{d}$$
and
$$\nabla^2 D(\bm{x})= \|\bm{x}\|^2\cdot \mathbb{I}+2\bm{x}\bm{x}^{\top}-\bm{Y}.$$
With direct computations,
$$\|\nabla^2 D(\bm{x})\|_{\textrm{op}}\leq 3\|\bm{x}\|^2+\|\bm{Y}\|_{\textrm{op}}.$$

1). As $\|\bm{x}\|\leq \sqrt{\|\bm{Y}\|_{\textrm{op}}/2}:=2a$,
$$\|\nabla^2 D(\bm{x})\|_{\textrm{op}}\leq 20a^2.$$

2). As  $\|\bm{x}\|\geq \sqrt{\|\bm{Y}\|_{\textrm{op}}/2}=2a$, we can get
$$\|\nabla D(\bm{x})\|=\|(\bm{x}\bm{x}^{\top}-\bm{Y})\bm{x}\|\geq \|\bm{x}\|^3-\|\bm{Y}\|_{\textrm{op}}\|\bm{x}\|\geq\frac{\|\bm{x}\|^3}{2}\geq a\|\bm{x}\|^2.$$
Thus, we get
$$\|\nabla^2 D(\bm{x})\|_{\textrm{op}}\leq \frac{3}{a}\|\nabla D(\bm{x})\|+8a^2.$$

In summary, we are led to
$$\|\nabla^2 D(\bm{x})\|_{\textrm{op}}\leq \frac{3}{a}\|\nabla D(\bm{x})\|+20a^2.$$
From [Corollary A.4, \cite{zhang2020improved}], we then proved that the function in \eqref{se} satisfies Assumption \ref{ass1}.

\medskip

\textbf{About Assumption \ref{ass2}:}
Noticing that $[\nabla^2 D(\bm{x})]_{:,:,k}=2\bm{x}_k \mathbb{I}+\bm{D}
$ with $\bm{D}_{i,j}=\frac{\partial(\bm{x}_i\bm{x}_j)}{\partial \bm{x}_k}$.

1. As $\|\bm{x}\|\leq 2a$,
$$\|[\nabla^3 D(\bm{x})]_{:,:,k}\|_F\leq 8da,$$
indicating
$$\|[\nabla^3 D(\bm{x})]\|_F\leq 8d^3a.$$

2. As  $\|\bm{x}\|\geq 2a$, it also holds
$$\|\nabla D(\bm{x})\|\geq a\|\bm{x}\|^2\geq 2a^2\|\bm{x}\|.$$
Thus, we get
$$\|[\nabla^3 D(\bm{x})]_{:,:,k}\|_F\leq 4d\|\bm{x}\|\leq \frac{d}{a^2}\|\nabla D(\bm{x})\|,$$
giving us
$$\|[\nabla^3 D(\bm{x})]\|_F\leq  \frac{d^3}{a^2}\|\nabla D(\bm{x})\|.$$
In summary, we get $\|[\nabla^3 D(\bm{x})]\|_F\leq 8d^3a+\frac{d^3}{a^2}\|\nabla D(\bm{x})\|$
From Proposition \ref{pro2}, we then prove Assumption \ref{ass2}.
\subsection{Proof of Proposition \ref{pro2}}
1. For any unit $\bm{z}\in\RR^d$, we have
\begin{equation*}
\|\sum_{k=1}^d[\nabla^3  f({\bm x})]_{i,j,k}\bm{z}\|_{\textrm{op}}=\lim_{t\rightarrow 0}\frac{\|\nabla^2 f({\bm x}+t\bm{z})-\nabla^2  f({\bm x})\|_{\textrm{op}}}{t}\leq H_1+H_2\|\nabla  f({\bm x})\|.
\end{equation*}
By setting $\bm{z}=\bm{e}_k$, we then get
$|[\nabla^3  f({\bm x})]_{i,j,k}|\leq\|[\nabla^3  f({\bm x})]_{:,:,k}\|_{\textrm{op}}\leq H_1+H_2\|\nabla  f({\bm x})\|$. Thus, we get
$$\|\nabla^3  f({\bm x})\|_{F}\leq d^{3/2}H_1+d^{3/2}H_2\|\nabla  f({\bm x})\|.$$

\medskip

\noindent 2. Given two points $\bm{x}$ and $\bm{x}^{\dag}$ such that $\|\bm{x}-\bm{x}^{\dag}\|\leq R$, by denoting $\gamma(s):=\bm{x}+s(\bm{x}^{\dag}-\bm{x})$,
$$\nabla^2 f(\gamma(s))=\int_{0}^s [\nabla^3 f(\gamma(\tau))](\bm{x}^{\dag}-\bm{x})d\tau +\nabla^2 f(\gamma(0)).$$
Hence, we can get
\begin{align*}
\|\nabla^2 f(\gamma(s))\|_{\textrm{op}}&\leq \|\bm{x}^{\dag}-\bm{x}\|\int_{0}^s \|\nabla^3 f(\gamma(\tau))\|_F d\tau +\|\nabla^2 f(\bm{x})\|_{\textrm{op}}\\
&\leq R\int_{0}^s \|\nabla^3 f(\gamma(\tau))\|_F d\tau +\|\nabla^2 f(\bm{x})\|_{\textrm{op}}\\
&\leq \hat{H}_1Rs+\hat{H}_2R\int_{0}^s \|\nabla f(\gamma(\tau))\| d\tau +\|\nabla^2 f(\bm{x})\|_{\textrm{op}}\\
&\leq \hat{H}_1Rs+\hat{H}_2R\int_{0}^s \Big\|\int_{0}^{\tau}\nabla^2 f(\gamma(\nu))(\bm{x}^{\dag}-\bm{x})d\nu+\nabla^2 f(\bm{x})\Big\| d\tau +\|\nabla^2 f(\bm{x})\|_{\textrm{op}}\\
&\leq \hat{H}_2R^2s\int_{0}^s \|\nabla^2 f(\gamma(\nu))\|_{\textrm{op}}d\nu +\hat{H}_1Rs+(\hat{H}_2Rs+1)\|\nabla^2 f(\bm{x})\|_{\textrm{op}}.
\end{align*}
Based on Lemma \ref{lemma-gro},
\begin{align*}
\|\nabla^2 f(\gamma(s))\|_{\textrm{op}}&\leq(\hat{H}_1Rs+(\hat{H}_2Rs+1)\|\nabla^2 f(\bm{x})\|_{\textrm{op}})\exp(\frac{\hat{H}_2R^2s^2}{2})\\
&\leq (\hat{H}_1R+(\hat{H}_2R+1)\|\nabla^2 f(\bm{x})\|_{\textrm{op}})\exp(\frac{\hat{H}_2R^2}{2})\\
&\leq (\hat{H}_1R+(\hat{H}_2R+1)L_1+(\hat{H}_2R+1)L_2\|\nabla f(\bm{x})\|)\exp(\frac{\hat{H}_2R^2}{2})\\
&=\mathcal{O}(1+\|\nabla f(\bm{x})\|).
\end{align*}
With direct computations,
\begin{align*}
\|\nabla^2 f(\bm{x})-\nabla^2 f(\bm{x}^{\dag})\|_{\textrm{op}}&=\|\int_{0}^1 [\nabla^3 f(\gamma(s))](\bm{x}^{\dag}-\bm{x})ds \|_{\textrm{op}}\\
&\leq \|\bm{x}^{\dag}-\bm{x}\|\int_{0}^1 \|\nabla^3 f(\gamma(s))\|_F ds\\
&\leq \|\bm{x}^{\dag}-\bm{x}\|\Big(\hat{H}_1s+\hat{H}_2\int_{0}^1 \|\nabla f(\gamma(s))\| ds\Big)\\
&\leq\|\bm{x}^{\dag}-\bm{x}\| \Big(\hat{H}_1s+\hat{H}_2\int_{0}^1 \Big\|\int_{0}^{s}\nabla^2 f(\gamma(\nu))(\bm{x}^{\dag}-\bm{x})d\nu+\nabla^2 f(\bm{x})\Big\| ds\Big)\\
&\leq \|\bm{x}^{\dag}-\bm{x}\| \Big(\hat{H}_1R+\hat{H}_2R^2\int_{0}^1 \|\nabla^2 f(\gamma(s))\|_{\textrm{op}}ds+\hat{H}_2R\|\nabla^2 f(\bm{x})\|_{\textrm{op}}\Big)\\
&=\mathcal{O}\Big((1+\|\nabla f(\bm{x})\|)\|\bm{x}^{\dag}-\bm{x}\|\Big).
\end{align*}
\subsection{Proof of Proposition \ref{pro-cond}}
When $\|\bm{w}^{\ddag}-\bm{w}^{\dag}\|=\gamma\leq r$,   Lemma \ref{lemma0} gives us
\begin{align*}
f(\bm{w}^{\ddag})-f(\bm{w}^{\dag})&\leq \langle\nabla f(\bm{w}^{\dag}),\bm{w}^{\ddag}-\bm{w}^{\dag}\rangle+\frac{L_1+L_2\|\nabla f(\bm{w}^{\dag})\|}{2}\|\bm{w}^{\ddag}-\bm{w}^{\dag}\|^2\\
&\leq-\gamma \langle\nabla f(\bm{w}^{\dag}),\Tc(\bm{m})\rangle+\frac{L_1+L_2\|\nabla f(\bm{w}^{\dag})\|}{2}\gamma^2\|\Tc(\bm{m})\|^2\\
&\overset{a)}{\leq}-\gamma l\|\bm{m}\|_{\diamond}+\gamma U\|\bm{m}-\nabla f(\bm{w}^{\dag})\|+\frac{L_1+L_2\|\nabla f(\bm{w}^{\dag})\|}{2}U^2\gamma^2\\
&\overset{b)}{\leq}-\gamma l\|\nabla f(\bm{w}^{\dag})\|_{\diamond}+(U+b)\gamma\|\bm{m}-\nabla f(\bm{w}^{\dag})\|+\frac{L_1+L_2/a\|\nabla f(\bm{w}^{\dag})\|_{\diamond}}{2}U^2\gamma^2\\
&\overset{c)}{\leq} -\frac{\gamma l}{2}\|\nabla f(\bm{w}^{\dag})\|_{\diamond}+(U+b)\gamma\|\bm{m}-\nabla f(\bm{w}^{\dag})\|+\frac{L_1}{2}U^2\gamma^2
\end{align*}
as $\gamma\leq\frac{al}{U^2L_2}$,
where $a)$ is due to the fact that
\begin{align*}
-\langle\nabla f(\bm{w}^{\dag}),\Tc(\bm{m})\rangle&=-\langle\bm{m},\Tc(\bm{m})\rangle-\langle\nabla f(\bm{w}^{\dag})-\bm{m},\Tc(\bm{m})\rangle\\
&\leq-l\|\bm{m}\|_{\diamond}+\langle\nabla f(\bm{w}^{\dag})-\bm{m},\Tc(\bm{m})\rangle\\
&\leq -l\|\bm{m}\|_{\diamond}+U\|\bm{m}-\nabla f(\bm{w}^{\dag})\|,
\end{align*}
and $b)$ is because
$$- \|\bm{m}\|_{\diamond}\leq -\|\nabla f(\bm{w}^{\dag})\|_{\diamond}+\|\nabla f(\bm{w}^{\dag})-\bm{m}\|_{\diamond}\leq -\|\nabla f(\bm{w}^{\dag})\|_{\diamond}+b\|\nabla f(\bm{w}^{\dag})-\bm{m}\|,$$
and $c)$ depends on $\gamma^2L_2U^2/a\leq \gamma l.$
\subsection{Proof of Proposition \ref{pro-conver}}
The proof is almost identical to the one of Theorem \ref{th4} with $\|\cdot\|_1$ being replaced by $\|\cdot\|_{\diamond}$.
\section{Proofs of Technical Lemmas}
\subsection{Proof of Lemma \ref{lemma0}}
1. When Assumption \ref{lemma0} holds, i.e., $\|\nabla f(\bm{y})-\nabla f(\bm{x})\|\leq (L_1+L_2\|\nabla  f({\bm x})\|)\|\bm{x}-\bm{y}\|$ for $\bm{x},\bm{y}\in\RR^d$ such that $\|\bm{x}-\bm{y}\|\leq r$, we have
\begin{align*}
& f(\bm{y})-f(\bm{x})-\langle\nabla f(\bm{x}),\bm{y}-\bm{x}\rangle\\
&\leq\Big|\int_{h=0}^{1}\langle\nabla f(\bm{x}+(\bm{y}-\bm{x})h)-\nabla f(\bm{x}), \bm{y}-\bm{x}\rangle dh \Big|\\
&\leq \int_{h=0}^{1}\|\nabla f(\bm{y}+(\bm{x}-\bm{y})h)-\nabla f(\bm{x})\|\cdot\|\bm{x}-\bm{y}\|dh\\
&\leq (L_1+L_2\|\nabla  f({\bm x})\|)\|\int_{h=0}^{1}(1-h)\|\bm{x}-\bm{y}\|^2dh\\
&=\frac{(L_1+L_2\|\nabla  f({\bm x})\|)}{2}\|\bm{x}-\bm{y}\|^2.
\end{align*}

\noindent 2. Noticing $\|\nabla^2 f(\bm{y})-\nabla^2 f(\bm{x})\|_{\textrm{op}}\leq (H_1+H_2\|\nabla  f({\bm x})\|)\|\bm{x}-\bm{y}\|$ for $\bm{x},\bm{y}\in\RR^d$ such that $\|\bm{x}-\bm{y}\|\leq R$ due to Assumption~\ref{ass2},  we can see
\begin{align*}
& \|\nabla f(\bm{y})-\nabla f(\bm{x})-[\nabla^2 f(\bm{x})](\bm{y}-\bm{x})\| \\
&=\|\int_{h=0}^{1}[\nabla^2 f(\bm{x}+(\bm{y}-\bm{x})h)-\nabla^2 f(\bm{x})](\bm{y}-\bm{x}) dh \|\\
&\leq \int_{h=0}^{1}\|\nabla^2 f(\bm{y}+(\bm{x}-\bm{y})h)-\nabla^2 f(\bm{x})\|_{\textrm{op}}\|\bm{x}-\bm{y}\|dh\\
&\leq (H_1+H_2\|\nabla  f({\bm x})\|)\|\int_{h=0}^{1}(1-h)\|\bm{x}-\bm{y}\|^2dh\\
&=\frac{(H_1+H_2\|\nabla  f({\bm x})\|)}{2}\|\bm{x}-\bm{y}\|^2,
\end{align*}
as $\|\bm{x}-\bm{y}\|\leq R$.
\subsection{Proof of Lemma \ref{lemma1-sign}}
As $\|\bm{w}^{\ddag}-\bm{w}^{\dag}\|=\gamma\leq r$,   Lemma \ref{lemma0} yields
\begin{align*}
f(\bm{w}^{\ddag})-f(\bm{w}^{\dag})&\leq \langle\nabla f(\bm{w}^{\dag}),\bm{w}^{\ddag}-\bm{w}^{\dag}\rangle+\frac{L_1+L_2\|\nabla f(\bm{w}^{\dag})\|}{2}\|\bm{w}^{\ddag}-\bm{w}^{\dag}\|^2\\
&\leq-\gamma \langle\nabla f(\bm{w}^{\dag}),\SI(\bm{m})\rangle+\frac{L_1+L_2\|\nabla f(\bm{w}^{\dag})\|}{2}\gamma^2d\\
&\overset{a)}{=}-\gamma \|\bm{m}\|_1+\gamma \langle\bm{m}-\nabla f(\bm{w}^{\dag}),\SI(\bm{m})\rangle+\frac{L_1+L_2\|\nabla f(\bm{w}^{\dag})\|}{2}\gamma^2d\\
&\overset{b)}{\leq}-\gamma \|\nabla f(\bm{w}^{\dag})\|_1+2\gamma\|\bm{m}-\nabla f(\bm{w}^{\dag})\|_1+\frac{L_1+L_2\|\nabla f(\bm{w}^{\dag})\|}{2}\gamma^2d\\
&\leq-\gamma \|\nabla f(\bm{w}^{\dag})\|_1+2\gamma\|\bm{m}-\nabla f(\bm{w}^{\dag})\|_1+\frac{L_1+L_2\|\nabla f(\bm{w}^{\dag})\|_1}{2}\gamma^2d\\
&\overset{c)}{\leq} -\frac{\gamma}{2}\|\nabla f(\bm{w}^{\dag})\|_1+2\gamma\sqrt{d}\|\bm{m}-\nabla f(\bm{w}^{\dag})\|+\frac{L_1}{2}\gamma^2d
\end{align*}
as $\gamma\leq\frac{1}{L_2d}$,
where $a)$ is due to the fact that
\begin{align*}
-\langle\nabla f(\bm{w}^{\dag}),\SI(\bm{m})\rangle&=-\langle\bm{m},\SI(\bm{m})\rangle-\langle\nabla f(\bm{w}^{\dag})-\bm{m},\SI(\bm{m})\rangle\\
&=-\|\bm{m}\|_1+\langle\bm{m}-\nabla f(\bm{w}^{\dag}),\SI(\bm{m})\rangle\\
&\leq -\|\bm{m}\|_1+\|\bm{m}-\nabla f(\bm{w}^{\dag})\|_1,
\end{align*}
and $b)$ is because
$$- \|\bm{m}\|_1\leq -\|\nabla f(\bm{w}^{\dag})\|_1+\|\nabla f(\bm{w}^{\dag})-\bm{m}\|_1,$$
and $c)$ depends on $\gamma^2L_2d\leq \gamma.$
\subsection{Proof of Lemma \ref{lemma2}}
With the scheme of the algorithm,
\begin{align*}
\EE_{\CC,\chi^t}\Big\|\sum_{i=1}^n \bar{\bm{g}}^t(i)/n-\nabla f(\bm{v}^t)\Big\|^2&=\EE_{\CC,\chi^t}\Big\|\sum_{i=1}^n \bar{\bm{g}}^t(i)/n-\sum_{i=1}^n \bm{g}^t(i)/n+\sum_{i=1}^n \bm{g}^t(i)/n-\nabla f(\bm{v}^t)\Big\|^2\\
&\leq2\EE_{\CC,\chi^t}\Big\|\sum_{i=1}^n \bar{\bm{g}}^t(i)/n-\sum_{i=1}^n \bm{g}^t(i)/n\|^2+2\EE_{\chi^t}\|\sum_{i=1}^n \bm{g}^t(i)/n-\nabla f(\bm{v}^t)\Big\|^2.
\end{align*}
With direct computations, we have
\begin{align}\label{lemm3-reuse1}
\begin{aligned}
\EE_{\CC,\chi^t}\Big\|\sum_{i=1}^n \bar{\bm{g}}^t(i)/n-\sum_{i=1}^n \bm{g}^t(i)/n\|^2&\leq\frac{1}{n}\sum_{i=1}^n\EE_{\CC,\chi^t}\|\bar{\bm{g}}^t(i)-\bm{g}^t(i)\|^2\\
&\leq\frac{(1-\delta)^u}{n}\sum_{i=1}^n\EE_{\chi^t}\|\bm{g}^t(i)\|^2\\
&=\frac{(1-\delta)^u}{n}\sum_{i=1}^n\Big[\EE_{\chi^t}\|\bm{g}^t(i)-\nabla f_{i}(\bm{v}^t)\|^2+\EE_{\chi^t}\|\nabla f_{i}(\bm{v}^t)\|^2\Big]\\
&\leq(1-\delta)^u\sigma^2+\frac{(1-\delta)^u}{n}\sum_{i=1}^n\EE_{\chi^t}\|\nabla f_{i}(\bm{v}^t)-\nabla f(\bm{v}^t)+\nabla f(\bm{v}^t)\|^2\\
&\leq (1-\delta)^u\sigma^2+\frac{(1-\delta)^u}{n}\sum_{i=1}^n(2\EE_{\chi^t}\|\nabla f_{i}(\bm{v}^t)-\nabla f(\bm{v}^t)\|^2+2\|\nabla f(\bm{v}^t)\|^2)\\
&\leq (\sigma^2+2\bar{\sigma}^2)(1-\delta)^u+2(1-\delta)^u\|\nabla f(\bm{v}^t)\|^2.
\end{aligned}
\end{align}
Under Assumption \ref{ass3},
\begin{align*}
\EE_{\chi^t}\|\sum_{i=1}^n\bm{g}^t(i)/n-\nabla f(\bm{v}^t)\|^2&=\EE_{\chi^t}\|\sum_{i=1}^n\bm{g}^t(i)/n-\nabla f_{i}(\bm{v}^t)/n\|^2\\
&=\frac{1}{n^2}\sum_{i=1}^n\EE_{\chi^t}\|\bm{g}^t(i)-\nabla f_{i}(\bm{v}^t)\|^2\leq\frac{\sigma^2}{n}.
\end{align*}
In summary, we can get
\begin{align*}
\EE_{\CC,\chi^t}\Big\|\sum_{i=1}^n \bar{\bm{g}}^t(i)/n-\nabla f(\bm{v}^t)\Big\|^2\leq (2\sigma^2+4\bar{\sigma}^2)(1-\delta)^u+\frac{2\sigma^2}{n}+4(1-\delta)^u\|\nabla f(\bm{v}^t)\|^2.
\end{align*}
\subsection{Proof of Lemma \ref{lemma3}}
With the scheme of the algorithm, as $k\neq j$,
\begin{align*}
&2\EE_{\CC,\chi^t}\langle\sum_{i=1}^n\bar{\bm{g}}^k(i)/n-\nabla f(\bm{v}^k),\sum_{i=1}^n\bar{\bm{g}}^j(i)/n-\nabla f(\bm{v}^j)\rangle\\
&=2\EE_{\CC,\chi^t}\langle\sum_{i=1}^n\bar{\bm{g}}^k(i)/n-\nabla f(\bm{v}^k),\sum_{i=1}^n\bar{\bm{g}}^j(i)/n-\sum_{i=1}^n\bm{g}^j(i)/n+\sum_{i=1}^n\bm{g}^j(i)/n-\nabla f(\bm{v}^j)\rangle\\
&=2\EE_{\CC,\chi^t}\langle\sum_{i=1}^n\bar{\bm{g}}^k(i)/n-\nabla f(\bm{v}^k),\sum_{i=1}^n\bar{\bm{g}}^j(i)/n-\sum_{i=1}^n\bm{g}^j(i)/n\rangle\\
&\qquad\qquad+\underbrace{2\EE_{\CC,\chi^t}\langle\sum_{i=1}^n\bar{\bm{g}}^k(i)/n-\nabla f(\bm{v}^k),\sum_{i=1}^n\bm{g}^j(i)/n-\nabla f(\bm{v}^j)\rangle}_{=0}\\
&=2\EE_{\CC,\chi^t}\langle\sum_{i=1}^n\bar{\bm{g}}^k(i)/n-\sum_{i=1}^n\bm{g}^k(i)/n+\sum_{i=1}^n\bm{g}^k(i)/n-\nabla f(\bm{v}^k),\sum_{i=1}^n\bar{\bm{g}}^j(i)/n-\sum_{i=1}^n\bm{g}^j(i)/n\rangle\\
&=2\EE_{\chi^t}\langle\sum_{i=1}^n\bar{\bm{g}}^k(i)/n-\sum_{i=1}^n\bm{g}^k(i)/n,\sum_{i=1}^n\bar{\bm{g}}^j(i)/n-\sum_{i=1}^n\bm{g}^j(i)/n\rangle\\
&\qquad\qquad+\underbrace{2\EE_{\CC,\chi^t}\langle\sum_{i=1}^n\bm{g}^k(i)/n-\nabla f(\bm{v}^k),\sum_{i=1}^n\bar{\bm{g}}^j(i)/n-\sum_{i=1}^n\bm{g}^j(i)/n\rangle}_{=0}\\
&\leq \EE_{\chi^t}\|\sum_{i=1}^n\bar{\bm{g}}^k(i)/n-\sum_{i=1}^n\bm{g}^k(i)/n\|^2+\EE_{\chi^t}\|\sum_{i=1}^n\bar{\bm{g}}^j(i)/n-\sum_{i=1}^n\bm{g}^j(i)/n\|^2,
\end{align*}
where we used $\EE_{\CC,\chi^t}(\sum_{i=1}^n\bm{g}^k(i)/n)=\nabla f(\bm{v}^k)$ yielding that
\begin{align*}
&\EE_{\CC,\chi^t}\langle\sum_{i=1}^n\bar{\bm{g}}^k(i)/n-\nabla f(\bm{v}^k),\sum_{i=1}^n\bm{g}^j(i)/n-\nabla f(\bm{v}^j)\rangle\\
&\qquad=\langle\EE_{\CC,\chi^t}[\sum_{i=1}^n\bar{\bm{g}}^k(i)/n-\nabla f(\bm{v}^k)],\EE_{\CC,\chi^t}[\sum_{i=1}^n\bm{g}^j(i)/n-\nabla f(\bm{v}^j)\rangle]=0
\end{align*}
and
\begin{align*}
&\EE_{\CC,\chi^t}\langle\sum_{i=1}^n\bm{g}^k(i)/n-\nabla f(\bm{v}^k),\sum_{i=1}^n\bar{\bm{g}}^j(i)/n-\sum_{i=1}^n\bm{g}^j(i)/n\rangle\\
&\qquad=\langle\EE_{\CC,\chi^t}[\sum_{i=1}^n\bm{g}^k(i)/n-\nabla f(\bm{v}^k)],\EE_{\CC,\chi^t}[\sum_{i=1}^n\bar{\bm{g}}^j(i)/n-\sum_{i=1}^n\bm{g}^j(i)/n]\rangle=0.
\end{align*}
 Using \eqref{lemm3-reuse1}, we get
\begin{align*}
&2\EE_{\CC,\chi^t}\langle\sum_{i=1}^n\bar{\bm{g}}^k(i)/n-\nabla F(\bm{v}^k),\sum_{i=1}^n\bar{\bm{g}}^j(i)/n-\nabla F(\bm{v}^j)\rangle\\
&\qquad\qquad\leq (2\sigma^2+4\bar{\sigma}^2)(1-\delta)^u+2(1-\delta)^u\|\nabla f(\bm{v}^k)\|^2+2(1-\delta)^u\|\nabla f(\bm{v}^j)\|^2.
\end{align*}
 \medskip

\section{Details of the Experiments in Sec. 2.1}\label{sec2.1-numer}
We follow the same setting as Appendix H of \cite{zhang2019gradient}, the major difference is another formulation of local gradient Hessian smoothness.
Given the training trajectory $\{\bm{w}^0,\bm{w}^1,\ldots,\bm{w}^T\}$, the local gradient Hessian smoothness at $\bm{w}^t$ is estimated as
$$\hat{H}(\bm{w}^t)=\max_{h\in\Big\{\delta,2\delta,\ldots,1\}}\{\frac{\|\nabla f(\bm{w}^t+h\bm{d}^t)+\nabla f(\bm{w}^t-h\bm{d}^t)-2\nabla f(\bm{w}^t)\|}{h^2\|\bm{d}^t\|^2}\Big\},$$
where $\delta$ is a small positive number, and  $\bm{d}^t:=\bm{w}^{t+1}-\bm{w}^t$.

\end{document}